\documentclass{article}

\usepackage{arxiv}

\usepackage[utf8]{inputenc} 
\usepackage[T1]{fontenc}    
\usepackage{hyperref}       
\usepackage{url}            
\usepackage{booktabs}       
\usepackage{amsfonts}       
\usepackage{nicefrac}       
\usepackage{microtype}      
\usepackage{lipsum}		
\usepackage{graphicx}
\usepackage{natbib}
\usepackage{doi}
\usepackage{graphicx}
\usepackage{amsmath,amssymb, bm}
\usepackage[noend]{algpseudocode}
\usepackage{algorithm}
\usepackage{multirow}
\usepackage{mathtools}
\usepackage{dsfont}
\usepackage{stfloats}
\usepackage[caption=false,font=normalsize,labelfont=sf,textfont=sf]{subfig}
\usepackage{booktabs}
\usepackage{pifont}
\usepackage{geometry}
\usepackage[inkscapelatex=false]{svg}
\usepackage{comment}
\usepackage{tikz}
\usetikzlibrary{arrows.meta, positioning, shapes.geometric, calc}
\newcommand{\cmark}{\ding{51}} 
\newcommand{\xmark}{\ding{55}} 
\usepackage{threeparttable} 
\usepackage{colortbl} 
\definecolor{selected}{rgb}{0.88,1,1}

\title{Bayesian Physics Informed Neural Networks for \\ Reliable Transformer Prognostics}

\author{%
  Ibai Ramirez$^{1}$ \quad 
  Jokin Alcibar$^{1}$ \quad 
  Joel Pino$^{1}$ \quad 
  Mikel Sanz$^{2,5,6}$ \quad 
  David Pardo$^{3,5,6}$ \quad 
  Jose I. Aizpurua$^{4,6}$\thanks{Corresponding author: \texttt{joxe.aizpurua@ehu.eus}} \\
  $^{1}$Department of Electronics \& Computing, Mondragon University \\
  $^{2}$Department of Physical Chemistry, University of the Basque Country (UPV/EHU) \\
  $^{3}$Department of Mathematics, University of the Basque Country (UPV/EHU) \\
  $^{4}$Department of Computer Science and Artificial Intelligence, University of the Basque Country (UPV/EHU) \\
  $^{5}$Basque Centre for Applied Mathematics (BCAM) \\
  $^{6}$Ikerbasque, Basque Foundation for Science
}

\renewcommand{\shorttitle}{\textit{arXiv} Template}

\hypersetup{
pdftitle={A template for the arxiv style},
pdfsubject={q-bio.NC, q-bio.QM},
pdfauthor={David S.~Hippocampus, Elias D.~Striatum},
pdfkeywords={First keyword, Second keyword, More},
}

\begin{document}
\maketitle

\begin{abstract}
	Scientific Machine Learning (SciML) integrates physics and data into the learning process, offering improved generalization compared with purely data-driven models. Despite its potential, applications of SciML in prognostics remain limited, partly due to the complexity of incorporating partial differential equations (PDEs) for ageing physics and the scarcity of robust uncertainty quantification methods. This work introduces a Bayesian Physics-Informed Neural Network (B-PINN) framework for probabilistic prognostics estimation. By embedding Bayesian Neural Networks into the PINN architecture, the proposed approach produces principled, uncertainty-aware predictions. The method is applied to a transformer ageing case study, where insulation degradation is primarily driven by thermal stress. The heat diffusion PDE is used as the physical residual, and different prior distributions are investigated to examine their impact on predictive posterior distributions and their ability to encode \textit{a priori} physical knowledge. The framework is validated against a finite element model developed and tested with real measurements from a solar power plant. Results, benchmarked against a dropout-PINN baseline, show that the proposed B-PINN delivers more reliable prognostic predictions by accurately quantifying predictive uncertainty. This capability is crucial for supporting robust and informed maintenance decision-making in critical power assets.
\end{abstract}


\section{Introduction}
\label{sec:Intro}

Scientific Machine Learning (SciML) is an emerging interdisciplinary field that integrates physics and data within the learning process \cite{karniadakis2021physicsinformed}. Several SciML solutions have been proposed to address scientific and engineering problems, including Physics-Informed Neural Networks (PINNs) \cite{Raissi_19}, Kolmogorov Arnold Networks (KANs) \cite{KAN_ICLR}, Neural operators \cite{NeuralOperators_23}, Physics-informed Neural Operators \cite{PINO_24}, and their variants \cite{toscano2025}.

In the area of Prognostics \& Health Management (PHM), hybrid prognostics models have been widely explored \cite{aizpurua2015towards, Guo_20, Zio_22},  which are conceptually aligned with SciML principles \cite{Xu_23}. These approaches integrate physics-of-failure models with machine learning (ML) methods through (i) sequential configurations, \textit{i.e.} connect physics model outcomes with ML models \cite{Chao22} or vice-versa \cite{daigle2015model}, and (ii) parallel configurations, where physics and ML models are used simultaneously and their outcomes are fused, e.g. for error-correction configurations \cite{Alcibar_2025}.

The focus of this work is on SciML approaches that learn the dynamics of the physics model along with the ML model. Namely, PINNs focus on the use of Neural Network (NN) architectures, and they learn to solve partial differential equations (PDE) through modifications in the NN loss function \cite{Raissi_19}. Despite their promise, PINN-based solutions for PHM remain limited \cite{Xu_23}. There have been solutions focused on battery prognostics through imposing a monotonic degradation constraint in the loss function \cite{wang2024physics} and transformer prognostics through heat diffusion PDEs \cite{Ramirez_25}.

Other relevant hybrid prognostics methods have focused on the integration of principled models in a Recurrent Neural Network for Li-Ion battery prognostics \cite{nascimento2023framework}. The approach captures epistemic uncertainty through variational layers to estimate reduced-order model parameters. Similarly, \cite{Juan_2023} proposed a physics-guided Bayesian Neural Network (BNN) for reinforced concrete columns under lateral loading, using Approximate Bayesian Computation for robust uncertainty quantification.

It can be observed that PINN-based solutions for PHM have been focused on deterministic predictions. This is aligned with the observation that, while the progress and application of ML and AI models is rapidly increasing in different scientific and engineering areas, in contrast, the development of ML reliability assessment methods are relatively underdeveloped. In the context of failure prognostics, uncertainty quantification (UQ) is essential for reliable future ageing predictions \cite{sankararaman2015significance}. To this end, different UQ methods have been developed, including state-space models incorporating Bayesian principles \cite{daigle2015model}, Gaussian Process based approaches \cite{biggio2021uncertainty}, Conformal Prediction for prognostics \cite{javanmardi2023conformal}, and Bayesian Convolutional NNs \cite{PHME_Jokin}. For a complete picture of UQ for prognostics, refer to an extensive tutorial \cite{nemani2023uncertainty} and benchmarking studies \cite{basora2025benchmark}.

Within the SciML community, extensions of PINNs with UQ have been explored by  replacing deterministic NNs with BNNs, leading to Bayesian-PINNs (B-PINNs) \cite{Linka2022}. More broadly, Bayesian deep learning integrated with physics principles has shown solid and reliable results, not only for PINNs \cite{flores2025}, but also for Neural Operators \cite{Garg2023} and KANs \cite{gao2025scalable}. Other principled UQ alternatives for PINNs have focused on the modification of the output layer with a Gaussian process \cite{Li_2024}. Refer to \cite{psaros2023uncertainty} for an extensive review. 

Additionally, there have been studies that focused on the development of wrapper models that operate as an add-on to existing PINN models for UQ, such as Epistemic-PINN (EPINN), which are trained through Hamiltonian Monte Carlo (HMC) methods \cite{Epinet2025}  or Conformal Prediction PINNs  (C-PINNs), which provides guaranteed uncertainty estimates \cite{podina2024conformalized}.

Simpler yet effective UQ modelling techniques include Monte Carlo dropout \cite{Gal16} and deep ensemble methods \cite{Lakshminarayanan_17}, also used for PINNs, \textit{i.e.} dropout-PINN (dPINN) \cite{Zhang_19} and deep ensemble PINNs (EnsPINN) \cite{jiang2023practical, zou2025learning}.

Table~\ref{tab:uq_comparison} summarizes representative SciML and hybrid PHM methods, highlighting if they incorporate UQ, adhere to SciML principles, their underlying method, and the specific PHM application considered (if addressed).

\begin{table}[!htb]
	\small
	\centering
	\setlength{\tabcolsep}{3.500pt}
	\caption{Comparison of SciML and hybrid PHM methods, including relevant features for this work.}
	\label{tab:uq_comparison}
	\begin{tabular}{lcccc}
		\toprule
		\textbf{Method} & \textbf{UQ} & \textbf{SciML} & \textbf{Method} & \textbf{PHM} \\
		\midrule
		\cite{Juan_2023}         & \cmark & \xmark & BNN & Material \\
		\cite{nascimento2023framework} & \cmark & \xmark & RNN & Battery \\
		\cite{wang2024physics}         & \xmark & \cmark & PINN & Battery \\
		\cite{Ramirez_25} & \xmark & \cmark & PINN & Transformer \\
		\cite{Linka2022} & \cmark & \cmark & B-PINN & \xmark \\
		\cite{podina2024conformalized}  & \cmark & \cmark & C-PINN & \xmark \\
		\cite{Epinet2025}         & \cmark & \cmark & EPINN & \xmark \\
		\cite{Zhang_19}  & \cmark & \cmark & dPINN & \xmark \\
		\cite{zou2025learning}  & \cmark & \cmark & EnsPINN & \xmark \\
		Ours     & \cmark & \cmark & B-PINN & Transformer \\
		\bottomrule
	\end{tabular}
\end{table}
\normalsize


In this context, it can be observed that, focused on prognostics activities, existing PINN models have been based on deterministic models. In the broader SciML community, uncertainty-aware PINN methods have been proposed, but to the best of authors' knowledge, tested on synthetic controlled problems without addressing PHM problems.

Accordingly, the main contribution of this work is the development of a Bayesian-PINN model for electrical transformer prognostics. The proposed approach predicts a full posterior distribution for each spatiotemporal coordinate, which enables the inference of the maximum likelihood estimate along with the confidence of the model on the prediction.

The influence of different sources of uncertainty is thoroughly investigated to determine the most appropriate hyperparameters. Furthermore, the robustness of the designed B-PINN model is evaluated with respect to noise of the input and residual data points. The methodology is validated on a real transformer case study using probabilistic evaluation metrics. Obtained results demonstrate that the proposed B-PINN (i) delivers more accurate results than the dropout-PINN, with the added flexibility of specifying prior knowledge, and (ii) produces more reliable predictions than vanilla PINN model due to the probabilistic predictions and inferred uncertainty levels.

The remainder of this article is organized as follows. Section~\ref{sec:BasicsTrafo} provides background theory to understand transformer thermal and ageing modelling. Section~\ref{sec:Approach} introduces the proposed approach. Section~\ref{sec:CaseStudy} presents the case study. Section \ref{sec:ResultsDiscussion} presents the results, and finally, Section~\ref{sec:Conclusions} concludes.

\section{Power Transformer Thermal and Ageing Modelling}
\label{sec:BasicsTrafo}


Transformers are key assets for the reliable operation of the power grid. The increasing penetration of renewable energy sources (RESs) to the grid affects the transformer's health \cite{Aizpurua_23}. The main insulating material for oil-immersed transformers is paper immersed oil, and their main failure mode is the insulation degradation. The insulating paper is made of cellulose polymer and the degree of polymerization determines the strength of the insulating paper \cite{IEC60076_transf12}. The insulation paper degradation is directly caused by the thermal stress. 

Accordingly, this section reviews the main transformer thermal modelling steps, including an oil temperature estimation stage (Section~\ref{ss:PDE}) and subsequent winding temperature estimation stage, which is used to calculate the hottest-spot temperature (HST), \textit{i.e.} highest insulation temperature (Section~\ref{ss:Winding}). Finally, the thermal model is connected to the insulation aging assessment model, which is used to estimate the loss of life of the insulation (Section~\ref{ss:AgeingEstimate}).

\subsection{Spatiotemporal Oil Temperature Modelling}
\label{ss:PDE}

The spatial distribution of oil and winding temperature is key for the cost-effective transformer health management. To capture this, a thermal modelling approach is developed  based on partial differential equations (PDE). The heat diffusion PDE is considered to model the temporal and spatial evolution of the transformer oil temperature ($\Theta_{O}(x,t)$).  Due to the transformer oil characteristics, radiative heat diffusion is considered and not convection. The general form of the one-dimensional heat diffusion equation is defined as \cite{Ramirez_25}:

\vspace{-1.5mm}

\small
\begin{equation}
	\label{eq:PDE_1D_Diffusion}
	\begin{split}
		k\frac{\partial^2\Theta_{O}(x,t)}{\partial x^2} + q(x,t) &= \rho c_p \frac{\partial \Theta_{O}(x,t)}{\partial t} \\
		\frac{\partial^2\Theta_{O}(x,t)}{\partial x^2} + \frac{1}{k}q(x,t) &= \frac{1}{\alpha} \frac{\partial \Theta_{O}(x,t)}{\partial t}
	\end{split}
\end{equation}
\normalsize

\vspace{-1.5mm}

\noindent where $x,t$\hspace{1.0 mm}$\in$\hspace{0.4mm}$\mathbb{R}$ are the independent variables, which denote position [m] and time [s], respectively, $\Theta_{O}(x,t)$ is given in Kelvin [K], $k$ is the thermal conductivity [W/m.K], $c_p$ is the specific heat capacity [J/kg.K], $\rho$ is the density [kg/m\textsuperscript{3}], $q(x,t)$ is the rate of heat generation [W/m\textsuperscript{3}], and $\alpha=\frac{k}{\rho c_p}$ is the thermal diffusivity [m\textsuperscript{2}/s].

\begin{figure}[!htb]
	\centering
	\includegraphics[width=.45\columnwidth]{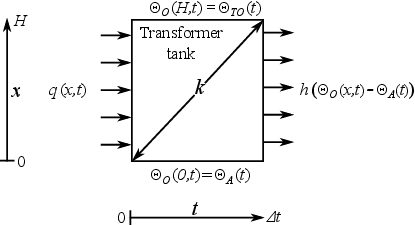}
	\caption{Transformer heat diffusion model.}
	\label{fig:Trafo_Diffusion_Basic}
\end{figure}

Figure~\ref{fig:Trafo_Diffusion_Basic} shows the thermal parameters of the transformer of the heat diffusion model  \cite{Ramirez_25}, which considers the heat source, $q(x,t)$, and the convective heat transfer, $h(\Theta_{O}(x,t)-\Theta_{A}(t))$.

\vspace{-1.5mm}

\vspace{-2 mm}

The evolution of the heat source in space and time, $q (x,t)$, is defined as follows:

\vspace{-1.5mm}

\small
\begin{equation}
	\label{eq:heat-source}
	q(x,t)=P_0+P_K(t)-h\left(\Theta_{O}(x,t)-\Theta_{A}(t)\right)
\end{equation}
\normalsize

\vspace{-1.5mm}

\noindent where $\Theta_{A}(t)$ is the ambient temperature, $h$ is the convective heat transfer coefficient, $P_0$ is the no load losses [W], and $P_K(t)$ is the load losses, defined as follows:

\vspace{-1.5mm}

\small
\begin{equation}
	P_K(t)=K(t)^2\mu,
\end{equation}
\normalsize

\vspace{-1.5mm}

\noindent where $K(t)$ is the load factor [p.u.], and $\mu$ is the rated load losses [W].

The challenge is to accurately model the transformer oil temperature vertically along its height $H$ (Figure~\ref{fig:Trafo_Diffusion_Basic}). It can be observed that the spatial distribution is considered along the vertical axis ($x$). It is assumed that $\Theta_{O}(x,t)$ is equal to $\Theta_{A}(t)$ at the bottom ($x$\hspace{0.5mm}=\hspace{0.5mm}$0$) and $\Theta_{TO}(t)$ at the top ($x$\hspace{0.5mm}=\hspace{0.5mm}$H$). Namely, these are the Dirichlet boundary conditions of the PDE that is going to be solved:

\vspace{-1.5mm}

\small
\begin{equation}
	\label{eq:BoundaryConditions}
	\begin{split}
		\Theta_{O}(0,t)&=\Theta_{A}(t)\\
		\Theta_{O}(H,t)&=\Theta_{TO}(t)\\
	\end{split}
\end{equation}
\normalsize

\vspace{-1.5mm}


\subsection{Hotspot Temperature (HST) Modelling}
\label{ss:Winding}

The HST is the most critical thermal indicator. However, direct measurements are difficult and expensive. Generally the HST value, $\Theta_{H}(t)$, is estimated indirectly from the top-oil temperature (TOT), $\Theta_{TO}(t)$, defined as \cite{IEC60076_transf12}:

\vspace{-1.5mm}

\small
\begin{equation}
	\label{eq:HST}
	\Theta_{H}(t)=\Theta_{TO}(t)+\Delta\Theta_{H}(t)
\end{equation}
\normalsize

\vspace{-1.5mm}

\noindent where $\Delta\Theta_{H}(t)$ is HST rise over TOT and $t$\hspace{0.4mm}$\in$\hspace{0.4mm}$\mathbb{R}$ is time.

With the spatiotemporal oil temperature estimate $\hat{\Theta}_{O}(x,t)$, the winding temperature distribution, $\hat{\Theta}_{W}(x,t)$, is defined as follows: 

\vspace{-2mm}
\small
\begin{equation}
	\label{eq:HST_spatial}
	\hat{\Theta}_{W}(x,t)={\hat{\Theta}_{O}}(x,t)+\Delta\Theta_{H}(t)
\end{equation}
\normalsize

\vspace{-1.5mm}

\noindent where, $x,t$\hspace{1.0 mm}$\in$\hspace{0.4mm}$\mathbb{R}$ are the position and time, respectively, $\hat{\Theta}_{O}(x,t)$ is the spatiotemporal oil temperature estimate. The hottest spatial temperature in Eq.~(\ref{eq:HST_spatial}) at each time instant $t$, corresponds to the HST in Eq.~(\ref{eq:HST}), \textit{i.e.} $max(\hat{\Theta}_{W}(\cdot,t))$=$\Theta_{H}(t)$.

$\Delta\Theta_{H}(t)$ is the HST rise over TOT, which is given by:

\vspace{-2 mm}
\small
\begin{equation}
	\label{eq:delta_HST}
	\Delta\Theta_{H}(t)=\Delta\Theta_{H_1}(t)-\Delta\Theta_{H_2}(t)
\end{equation}
\normalsize

\vspace{-2 mm}
\noindent where $\Delta\Theta_{H_1}(t)$ and $\Delta\Theta_{H_2}(t)$ model the oil heating considering the HST variations defined as follows \cite{IEC60076_transf12}:

\vspace{-2 mm}
\small
\begin{equation}
	\label{eq:HST_Transient1_1}
	d\Delta\Theta_{H_i}(t)=\upsilon_i\left[\beta_iK(t)^y-\Delta\Theta_{H_i}(t)\right]
\end{equation}
\normalsize

\vspace{-2 mm}

\noindent where {\small$K(t)$} is the load factor {\small[p.u.]}, $y$ is the winding exponent constant, which models the loading exponential power with the heating of the windings, {\small$i$\hspace{0.3mm}=\hspace{0.3mm}\{1,\hspace{0.4mm}2\}}, {\small$\upsilon_1$\hspace{0.2mm}=\hspace{0.3mm}$\Delta t/k_{22}\tau_w$}, {\small$\beta_1$\hspace{0.2mm}=\hspace{0.3mm}{$k_{21}\Delta\Theta_{H,R}$}} both for {\small$i$\hspace{0.3mm}=\hspace{0.3mm}$1$}, and {\small$\upsilon_2$\hspace{0.2mm}=\hspace{0.3mm}{$k_{22}\Delta t/\tau_{TO}$}}, {\small$\beta_2$\hspace{0.2mm}=\hspace{0.3mm}$(k_{21}$\hspace{0.3mm}-\hspace{0.3mm}$1)${$\Delta\Theta_{H,R}$}} {both} for {\small$i$\hspace{0.3mm}=\hspace{0.3mm}$2$}. {\small$\Delta t$}\hspace{0.3mm}{\small=}\hspace{0.3mm}$t$\hspace{0.5mm}-\hspace{0.5mm}$t${\small$^\prime$}, {\small$\tau_w$} and {\small$\tau_{TO}$} are the winding and oil time constants, {\small$k_{21}$} and {\small$k_{22}$} are the transformer thermal constants, and {\small$\Delta\Theta_{H,R}$} is the HST rise at rated load. The operator {\small$d$} denotes a difference operation on $\Delta t$, such that {\small$d\Delta\Theta_{H_i}(t)$\hspace{0.3mm}=\hspace{0.3mm}$\Delta\Theta_{H_i}(t)$\hspace{0.3mm}-\hspace{0.3mm}$\Delta\Theta_{H_i}(t^\prime)$} also for {\small$i$\hspace{0.3mm}=\hspace{0.3mm}\{1,\hspace{0.4mm}2\}}. To guarantee numerical stability, {\small$\Delta t$} should be small, never greater than half of the smaller time constant. 

Under steady state, the initial condition, {\small$\Theta_{H}(0)$}, can be defined as:

\vspace{-1.5mm}
\small
\begin{equation}
	\label{eq:InitCond_HST}
	\Theta_{H}(0)\!=\!\Theta_{TO}(0)\!+\!k_{21}\Delta\Theta_{H,R}K(0)^y\!-\!(k_{21}\!-\!1)\Delta\Theta_{H,R}K(0)^y
\end{equation}
\normalsize

\vspace{-1.5mm}

Eq.~(\ref{eq:InitCond_HST}) allows iteratively estimating the next HST values, $\Theta_{H}(n\Delta t)$, $n\in\mathbb{Z}^+$, using Eqs.~(\ref{eq:HST}), (\ref{eq:delta_HST}), and (\ref{eq:HST_Transient1_1}).

\subsection{Ageing Assessment}
\label{ss:AgeingEstimate}

The IEC 60076-7 standard defines insulation ageing acceleration factor at time $t$, {\small$V(t)$}, as \cite{IEC60076_transf12}:

\vspace{-1.5mm}

\small
\begin{equation}
	\label{eq:ageing_factor}
	V(t)=2^{\left(\Theta_{H}(t)-98\right)/6}
\end{equation}
\normalsize

\vspace{-1.5mm}

Using the spatiotemporal winding temperature $\hat{\Theta}_{W}(x,t)$, the ageing acceleration factor at time $t$ and position $x$, {\small$V(x,t)$}, can be defined as:

\vspace{-1.5mm}

\small
\begin{equation}
	\label{eq:ageing_factor_spatial}
	V(x,t)=2^{\left(\hat\Theta_W(x,t)-98\right)/6}
\end{equation}
\normalsize

\vspace{-1.5mm}

The IEC 60076-7 assumes an expected life of 30 years, with a reference HST of 98$^\circ$C \cite{IEC60076_transf12}. The loss of life (LOL) at location $x$ and time $t$ can be defined as:

\vspace{-1.5mm}

\small
\begin{equation}
	\label{eq:LoL}
	LOL(x,t)=\int_0^t V(x,t)dt
\end{equation}
\normalsize

\vspace{-1.5mm}

Consequently, the LOL at discrete time $L\Delta t$ and location $x$ can be obtained by summing the ageing (cf. Eq.~(\ref{eq:ageing_factor_spatial})) evaluated at the same time instants:

\vspace{-2mm}
\small
\begin{equation}
	\label{eq:lifetime_spatial}
	LOL(x,L\Delta t)=\sum_{n=0}^{L} V(x,n\Delta t)
\end{equation}
\normalsize

\vspace{-2 mm}

\noindent where  $n,L$\hspace{1.0 mm}$\in$\hspace{0.4mm}$\mathbb{Z}$\hspace{0.1mm}$^+$.

LOL can be converted into a recurrence relation for remaining useful life estimation \cite{Aizpurua_23}.

\section{Probabilistic Ageing Approach based on Bayesian PINNs}
\label{sec:Approach}

\begin{figure*}[!htb]
	\centering
	\includegraphics[width=.7\columnwidth]{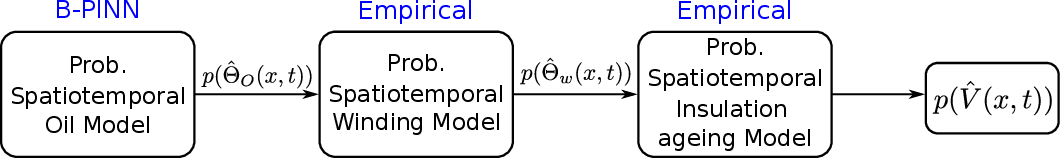}
	\caption{Overall framework for probabilistic transformer insulation ageing estimation. The Bayesian PINN-based oil temperature model is coupled with an empirical winding temperature model and an empirical insulation ageing model.}
	\label{fig:Framework_General}
\end{figure*}

The proposed transformer ageing estimation approach is based on a probabilistic spatiotemporal thermal stress model coupled with an ageing estimation model, as illustrated in Figure~\ref{fig:Framework_General}.  The thermal model is formulated using a Bayesian-PINN, which estimates the spatiotemporal oil temperature $\hat{\Theta}_o(x,t)$. This estimate is fed into an empirical winding temperature estimation model, that computes the spatiotemporal winding temperature, $\hat{\Theta}_w(x,t)$. Finally, the winding temperature is connected to an empirical insulation ageing model, which calculates the probabilistic ageing estimate, $p(\hat{V}(x,t))$.

\begin{figure*}[!htb]
	\centering
	\includegraphics[width=0.95\linewidth]{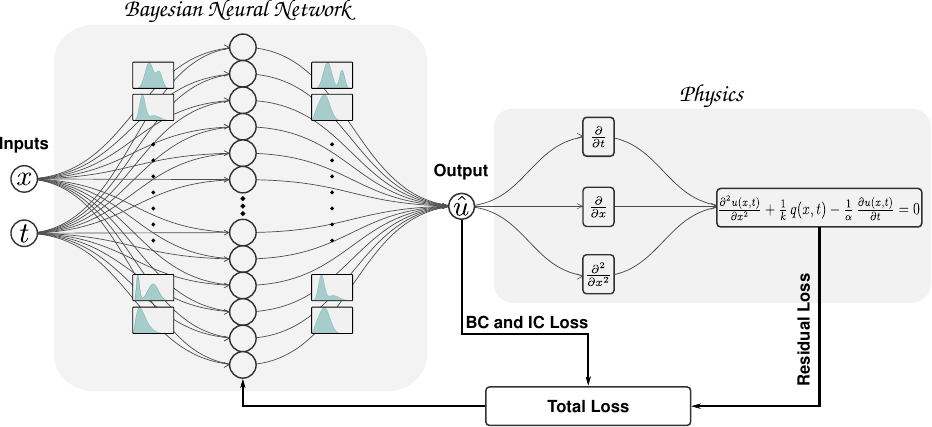}
	\caption{Proposed Bayesian-PINN approach for the probabilistic spatiotemporal transformer thermal model.}
	\label{fig:BPINN_Framework_General}
\end{figure*}

For clarity, this section focuses on the B-PINN approach. Figure~\ref{fig:BPINN_Framework_General} shows the proposed method, which integrates PINNs with Bayesian inference by replacing the deterministic NNs in PINNs with BNNs. Accordingly, Section~\ref{ss:PINN_Basics} introduces PINN basics, Section~\ref{ss:BNN_Basics} defines BNNs, and Section~\ref{ss:B-PINN} describes the B-PINN approach.

\vspace{-2 mm}
\subsection{PINN basics}
\label{ss:PINN_Basics}

PINNs were introduced with the goal of encoding PDE-based physics in ML models \cite{Raisi_19}, taking advantage of the ability of NNs  to act as universal approximators \cite{wright2022deep}. A general PDE can be written as \cite{Agnostopoulos_RBA}:

\vspace{-1 mm}

\small
\begin{equation}
	\label{eq:PDE_generic}
	d\{u(x,t)\}=f(x,t)
\end{equation}
\normalsize

\vspace{-1 mm}

\noindent where $x,t$$\in$$\mathbb{R}$ denote position and time, $u(x,t)$ is the unknown solution, $d$ is the differential operator, and $f(x,t)$ is a forcing function introducing external influences.

PINNs consist of the NN part, in which the inputs define temporal ($t$\hspace{0.4mm}$\in$\hspace{0.4mm}$\mathbb{R}$) and spatial coordinates ($x$\hspace{0.4mm}$\in$\hspace{0.4mm}$\mathbb{R}$ for one-dimensional cases) for the initial conditions (IC) and boundary conditions (BC). The NN output is an approximated PDE solution at the space and time coordinates, denoted {$\hat{u}(x,t)$}. This is calculated through the iterative application of weights (\bm{$w$}), biases (\bm{$b$}), and non-linear activation functions ($\sigma$) over the input. Namely, the inputs are connected through neurons, where they are multiplied with the weights and summed with the bias term. Finally, the weighted sum is passed through an activation function ($\sigma$). Subsequently, the outcome {$\hat{u}(x,t)$} is post-processed via automatic differentiation to compute the derivatives in space and time at certain collocation points (CP), generated via random sampling in the interior of the domain. The PINN is trained at these CPs by minimizing the residuals of the underlying PDE. 

The loss function, $\mathcal{L}(\bm{\theta},\lambda_0,\lambda_b,\lambda_r)$, incorporates the prediction error of the NN at IC and BC, and the residual of the PDE estimated via automatic differentiation at CP:

\vspace{-1 mm}
\small 
\begin{equation}
	\label{eq:Loss_PINN_generic}
	\mathcal{L}(\bm{\theta},\lambda_0,\lambda_b,\lambda_r)=\mathcal{L}_0(\bm{\theta},\lambda_0)+\mathcal{L}_b(\bm{\theta},\lambda_b)+\mathcal{L}_r(\bm{\theta},\lambda_r)
\end{equation}
\normalsize

\vspace{-1 mm}

\noindent where {$\bm{\theta}$\hspace{0.5mm}=\hspace{0.5mm}$\{\bm{w,b}\}$} are the weights and bias terms of the NN, and $\mathcal{L}_0(\bm{\theta},\lambda_0),\mathcal{L}_b(\bm{\theta},\lambda_b)$, and $\mathcal{L}_r(\bm{\theta},\lambda_r)$ are, respectively, the loss terms corresponding to IC, BC, and the residual of the PDE with their corresponding weights, $\lambda_0$, $\lambda_b$, and $\lambda_r$, defined as follows:

\vspace{-1.5 mm}
\small 
\begin{equation}
	\label{eq:Loss_IC}
	\mathcal{L}_0(\bm{\theta},\lambda_0)=\lambda_0\frac{1}{N_{0}}\sum_{i=1}^{N_{0}}|\hat{u}(x_i,0)-u(x_i,0)|^2
\end{equation}
\normalsize

\vspace{-1.5 mm}
\small 
\begin{equation}
	\label{eq:Loss_BC}
	\mathcal{L}_b(\bm{\theta},\lambda_b)=\lambda_b\frac{1}{N_{b}}\sum_{i=1}^{N_{b}}|\hat{u}(x_i,t_i)-u(x_i,t_i)|^2
\end{equation}
\normalsize

\vspace{-1.5 mm}
\small 
\begin{equation}
	\label{eq:Loss_r}
	\mathcal{L}_r(\bm{\theta},\lambda_r)=\lambda_r\frac{1}{N_r}\sum_{i=1}^{N_r}|(r(x_i,t_i)|^2
\end{equation}
\normalsize

\vspace{-1.5 mm}

\noindent where {$N_0,N_b$}, and {$N_r$} are, respectively, the number of IC, BC, and residue points, while {$u(x_i,t_i)$} and {$r(x_i,t_i)$} denote the known solution and the residual of PDE, for each training point $i$ defined at the coordinates {\small$(x_i,t_i)$}. According to the PDE defined in Eq.~{(\ref{eq:PDE_generic}), the residual {$r(x,t)$} is defined as follows:} 

\vspace{-1.5 mm}

\small
\begin{equation}
	\label{eq:Residual_generico}
	r(x,t)=d\{u(x,t)\}-f(x,t)
\end{equation}
\normalsize

\vspace{-1.5 mm}

Unlike classical NNs, which only minimize the prediction error at measurement points, PINNs enforce governing physics throughout the domain.

Minimizing the loss function in Eq.~{(\ref{eq:Loss_PINN_generic})} using a suitable optimization algorithm provides an optimal set of NN parameters $\bm{\theta}$\hspace{0.3mm}=\hspace{0.3mm}$\{\bm{w,b}\}$. That is, approximating the PDE becomes equivalent to finding $\bm{\theta}$ values that minimize the loss with a predefined accuracy. Overall, the key training parameters include: number of neurons and number of layers, number of CP, activation function, and the optimizer. Finding the correct solution requires knowing IC and BC. Additionally, random locations $(x_i,t_i)$, named CP, are used to evaluate the residual loss in Eq.~(\ref{eq:Loss_r}). 

The main motivation for using PINNs over numerical methods is the computational effort and adaptability to different solutions. Numerical models require a mesh of parameters to model and evaluate the PDE. As for PINNs, there is no need to define the whole mesh.

\vspace{-1 mm}

\subsection{Bayesian Neural Networks basics}
\label{ss:BNN_Basics}

Bayesian Neural Networks (BNNs) aim to estimate the posterior parameter distribution $P(\bm{\theta}|\mathcal{D})$, from a training dataset $\mathcal{D}=\{\bm{x}^{(i)},\bm{y}^{(i)}\}$, of a set of NN parameters $\bm{\theta}$:

\vspace{-1 mm}

\begin{equation}
	P(\bm{\theta}|\mathcal{D})=\frac{P(\mathcal{D}|\bm{\theta})P(\bm{\theta})}{P(\mathcal{D})}
	\label{eq:Bayes}
\end{equation}

\vspace{-1 mm}

\noindent where $P(\mathcal{D}|\bm{\theta})$ is the likelihood, $P(\bm{\theta})$ is the prior, \textit{i.e.} prior NN parameters knowledge expressed as a PDF over $\bm{\theta}$, and $P(\mathcal{D})$ is the marginal likelihood.

The likelihood is often estimated the individual product of pointwise estimated likelihoods $p_i(\mathcal{D}|\bm{\theta})$ based on the normal distribution $N(\mu,\sigma)$ defined as follows:

\vspace{-1 mm}

\begin{equation}
	P(\mathcal{D}|\bm{\theta})=\prod_{i=0}^{N}  p(y^{(i)}|x^{(i)},\bm{\theta})
\end{equation}

\vspace{-1 mm}

\begin{equation}
	\label{eq:Gaussian}
	p(y^{(i)}|x^{(i)},\bm{\theta})=\frac{1}{\sqrt{2\pi}\sigma}exp(-\frac{||x_i-x(t_i)||^2}{2\sigma^2})
\end{equation}

\vspace{-1 mm}

The selection of a good prior distribution, $P(\bm{\theta})$, for BNNs is challenging \cite{Fortuin_21}. The most common choice is to use a non-informative prior, which follows a Gaussian distribution with mean zero and unit variance, $\mathcal{N}(0,1)$, \textit{i.e.} isotropic Gaussian prior. In addition to the isotropic Gaussian prior, alternative priors implemented in this work include \textit{Spike-and-Slab} (SS) prior \cite{williams1995}, which is defined as follows:

\vspace{-1.5 mm}

\begin{equation}
	\label{eq:spike_slab_prior}
	P(\bm{\theta})_{\text{SS}} = \prod_{i=1}^d \left[ \pi \, \mathcal{N}(\theta_i; 0, \sigma_1^2) + (1 - \pi) \, \mathcal{N}(\theta_i; 0, \sigma_2^2) \right]
\end{equation}

\vspace{-1.5 mm}

\noindent where $\pi \in (0, 1)$ is the mixing coefficient, $\sigma_1$ and $\sigma_2$ correspond to the variance of the two Gaussian distributions, with $\sigma_1^2 \gg \sigma_2^2$, and $d$ is the dimensionality of $\bm{\theta}$. The first component (\textit{slab}) is a wide Gaussian that allows for larger parameter values, while the second (\textit{spike}) is a narrow Gaussian centered at zero, encouraging sparsity. This prior combines flexibility with regularization by allowing most weights to be close to zero while allowing a few to have large magnitudes.

In addition, the \textit{Laplace prior} will be also implemented \cite{williams1995}, defined as follows:

\vspace{-1 mm}

\begin{equation}
	\label{eq:L_prior}
	P(\bm{\theta})_{Laplace} = \prod_{i=1}^{d} \frac{\lambda}{2} \exp\left(-\lambda |\theta_i|\right)
\end{equation}

\vspace{-1 mm}

\noindent where $\lambda > 0$ is the scale parameter, and $d$ is the dimensionality of $\bm{\theta}$. The Laplace prior also promotes sparsity due to its sharp peak at zero and heavier tails compared to the non-informative Gaussian prior.

The analytical posterior in BNNs, $p(\bm{\theta}|\mathcal{D})$, is intractable, and therefore approximation methods are required. In this work, variational inference (VI) is used to approximate the true posterior with a variational distribution $q(\bm{\theta}|\bm{\phi})$ of known functional form by minimizing the Kullback–Leibler (KL) divergence between them. Since the true posterior cannot be computed directly, this minimization is performed indirectly by maximizing the Evidence Lower Bound (ELBO), which includes a KL divergence term between the variational distribution and the prior $p(\bm{\theta})$. The corresponding optimization objective is defined as follows:

\vspace{-1 mm}

\begin{equation}
	\mathcal{L}(\mathcal{D}, \bm{\phi}) = \text{KL}[q(\bm{\theta} \mid \bm{\phi}) \parallel p(\bm{\theta})] - \mathbb{E}_{q(\bm{\theta} \mid \bm{\phi})}[\log p(\mathcal{D} \mid \bm{\theta})]
\end{equation}

\vspace{-1 mm}

The first term is the KL divergence between the variational distribution $q(\bm{\theta}|\bm{\phi})$ and the prior $p(\bm{\theta})$ (known as the complexity cost). The second term is the expected value of the likelihood with respect to the variational distribution (known as the likelihood cost). The cost function can be also written as:

\vspace{-1 mm}

\small
\begin{equation}
	\begin{split}
		\label{eq:VI_log}
		\mathcal{L}(\mathcal{D}, \bm{\phi}) = 
		\mathbb{E}_{q(\bm{\theta}|\bm{\phi})} \left[ \log q(\bm{\theta}|\bm{\phi}) \right] 
		- \mathbb{E}_{q(\bm{\theta}|\bm{\phi})} \left[ \log p(\bm{\theta}) \right] 
		- \mathbb{E}_{q(\bm{\theta}|\bm{\phi})} \left[ \log p(\mathcal{D}|\bm{\theta}) \right]
	\end{split}
\end{equation}
\normalsize

\vspace{-1 mm}

It can be observed from Eq.~(\ref{eq:VI_log}) that loss terms are expectations with respect to the variational distribution $q(\bm{\theta}|\bm{\phi})$. Therefore, the cost function can be approximated by Monte Carlo sampling, drawing $N$ samples $\bm{\theta}^{(i)}$ from $q(\bm{\theta}|\bm{\phi})$:

\vspace{-1 mm}

\small
\begin{equation}
	\mathcal{L}(\mathcal{D},\!\bm{\phi})\! \approx \!
	\frac{1}{N}\! \sum_{i=1}^{N}\!\big[
	\log q(\bm{\theta}^{(i)}|\bm{\phi}) 
	\!-\! \log p(\bm{\theta}^{(i)}) 
	\!-\! \log p(\mathcal{D}|\bm{\theta}^{(i)}) \big]
	\label{eq:MC_ELBO}
\end{equation}
\normalsize

\vspace{-1 mm}

Throughout this paper, the variational posterior is assumed to follow a Gaussian distribution $\bm{\phi}=(\bm{\mu},\bm{\sigma})$, where $\bm{\mu}$ is the mean vector of the distribution and $\bm{\sigma}$ is the standard deviation vector.

\subsection{Bayesian PINNs}
\label{ss:B-PINN}


Bayesian PINNs (B-PINNs) extend standard PINNs by replacing the deterministic NN with a Bayesian NN. This enables the estimation of epistemic uncertainty, which reflects the model’s lack of complete knowledge~\cite{kendall2017}. In practice, epistemic uncertainty is quantified by placing probability density functions (PDFs) over the BNN parameters.

The main outcome of interest in the B-PINN approach is the posterior distribution. This is calculated following the Bayes rule (cf. Eq. ~(\ref{eq:Bayes})). Namely, the prior distribution is assumed to be given based on \textit{a priori} knowledge (\textit{i.e.}, Gaussian, Spike-and-Slab, or Laplace). The likelihood, $P(\mathcal{D}|\Theta)$, is estimated for the initial point, boundary condition points, and residual samples assuming a Gaussian distribution (cf. Eq.~(\ref{eq:Gaussian})), defined as follows:

\vspace{-1mm}
\small
\begin{align}
	\begin{split}
		& p(u_0 | x_0, \bm{\theta}^{(i)}) = \prod_{i=1}^{N_0} \frac{1}{\sqrt{2\pi} \sigma_0}\exp\left(\!-\frac{\| \hat{u}(x_0, 0; \bm{\theta}^{(i)}) - u_0 \|^2}{2 \sigma_0^2}\right)\\
		& p(\!u_{bc} | x_{bc}, t_{bc}, \bm{\theta}^{(i)}\!) \!=\! \prod_{i=1}^{N_{bc}}\! \frac{1}{\sqrt{2\pi} \sigma_{bc}}\!\exp\!\left(\!-\frac{\| \!\hat{u}(\!x_{bc}, t_{bc}; \bm{\theta}^{(i)}\!) \!- \!u_{bc}\! \|^2}{2 \sigma_{bc}^2}\!\right)\\
		& p(r(x_f, t_f; \bm{\theta}^{(i)})) = \prod_{i=1}^{N_{f}} \frac{1}{\sqrt{2\pi} \sigma_{f}}\exp\left(\!-\frac{\| r(x_f, t_f; \bm{\theta}^{(i)}) \|^2}{2 \sigma_{f}^2}\right)
	\end{split}
	\label{eq:L_BPINN}
\end{align}
\normalsize

\vspace{-1mm}

\noindent where $\sigma_0$, $\sigma_{bc}$, and $\sigma_f$ denote the standard deviation of initial, boundary, and residual points.

For computational tractability and efficiency, the log-likelihood terms are considered by taking the logarithm of Eq. (\ref{eq:L_BPINN}). Subsequently, the Monte Carlo ELBO loss defined for a generic BNN [cf. Eq.~(\ref{eq:MC_ELBO})], is adapted for B-PINN posterior inference:

\vspace{-1mm}

\small
\begin{equation}
		\mathcal{L}^{(i)} = 
		\log q(\bm{\theta}^{(i)}|\bm{w}) 
		- \log p(\bm{\theta}^{(i)}) 
		- \lambda_0\log p(u_0 | x_0, \bm{\theta}^{(i)}) 
		- \lambda_b\log p(u_{bc} | x_{bc}, t_{bc}, \bm{\theta}^{(i)})
		-\lambda_r \log p(r | x_f, t_f, \bm{\theta}^{(i)})
	\label{eq:ELBO_BPINN}
\end{equation}
\normalsize

\vspace{-1mm}

This process results in the approximation of the variational posterior distribution  $q(\bm{\theta}|\bm{w})$. The training process of the B-PINN approach is shown in Algorithm~\ref{alg:Bayesian_PINN}.


\begin{algorithm}[!htb]
	\caption{B-PINN Training via Variational Inference}
	\label{alg:Bayesian_PINN}
	\begin{algorithmic}[1]
		\State \textbf{Input:} Collocation points $(x_f, t_f)$, initial condition data $(x_0, u_0)$, boundary condition data $(x_{bc}, t_{bc}, u_{bc})$, prior distribution $p(\bm{\theta})$
		\State Initialize variational parameters $\bm{w} = \{\bm{\mu}, \bm{\sigma}\}$ for $\bm{\theta}$
		\While{not converged}
		\State Sample $\bm{\theta}^{(i)} \sim q(\bm{\theta}|\bm{w})$ via reparameterization trick
		\State Evaluate predicted solution: $\hat{u}(x,t;\bm{\theta}^{(i)})$
		\State Compute residual: $r(x_f,t_f;\bm{\theta}^{(i)}) = \mathcal{N}[\hat{u}](x_f, t_f;\bm{\theta}^{(i)}) - f(x_f,t_f)$
		\State Evaluate log-likelihood terms taking the $\log$ of Eq. (\ref{eq:L_BPINN})
		\State Evaluate prior log-probability: $\log p(\bm{\theta}^{(i)})$
		\State Evaluate variational density: $\log q(\bm{\theta}^{(i)}|\bm{w})$
		\State Compute Monte Carlo estimate of ELBO loss via Eq. (\ref{eq:ELBO_BPINN})
		\State Update $\bm{w} = \{\bm{\mu}, \bm{\sigma}\}$ using gradient $\nabla_{\bm{w}} \mathcal{L}^{(i)}$
		\EndWhile
		\State \textbf{Return:} Learned variational posterior $q(\bm{\theta}|\bm{w})$
	\end{algorithmic}
\end{algorithm}

\section{Case Study \& Experiments}
\label{sec:CaseStudy}

The case study examines the lifetime of a distribution transformer installed in a floating solar power plant in Spain \cite{Aizpurua_23}. The transformer's nameplate parameters are summarized in Table~\ref{table:CaseStudy_trafo}.

\begin{table}[!hbtp]
	\centering
	\caption{Transformer parameter values.}
	\label{table:CaseStudy_trafo}
	\begin{tabular}{|c|c|}
		\hline
		\small{\textbf{Parameter}} & \small{\textbf{Value}}\\ \hline	
		\small{Rating [kVA], V\textsubscript{1}/V\textsubscript{2}} & \small{1100, 22000/400}\\ 
		\small{R=Load losses/No load losses [W]} & \small{9800/842}\\ 		
		\small{$\Delta\Theta_{H,R}$} [$^\circ$C] & \small{15.1}\\ 	
		\small{$k_{21}$, $k_{22}$} & \small{2.32, 2.05}\\ 	
		\small{$\tau_0$, $\tau_w$ [min.]} & \small{266.8, 9.75} \\ 	\hline	
	\end{tabular}
	\vspace{-0.2cm}
\end{table}

\begin{figure}[!htb]
	\vspace{-0.2cm}
	\centering
	\includegraphics[width=0.45\columnwidth]{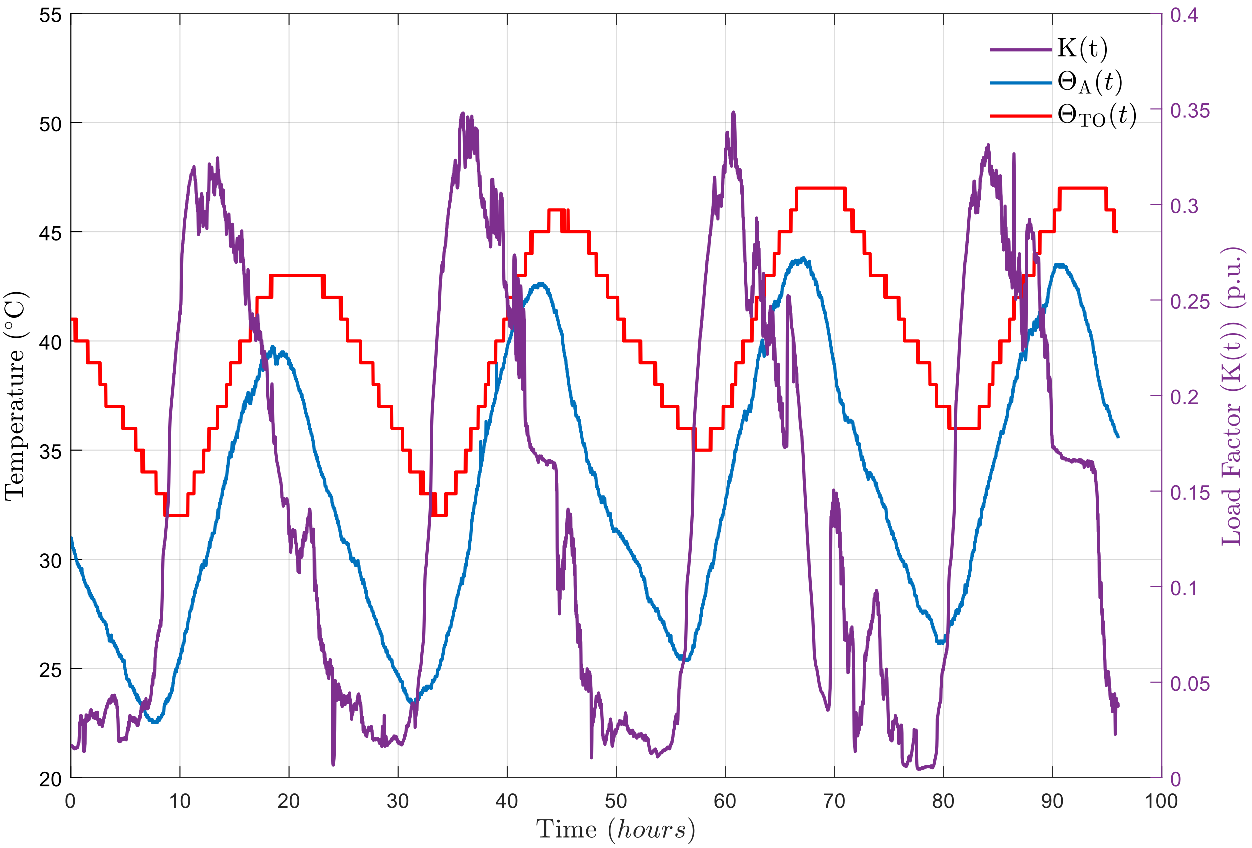}
	\vspace*{-0.2cm}
	\caption{Available load ($K$), ambient temperature ($\Theta_A$), and top-oil temperature ($\Theta_{TO}$) over four days of operation with one-minute resolution.}
	\label{fig:AvailableTimeSeries}
	\vspace{-0.3cm}
\end{figure}

The dataset was preprocessed by removing variables without measurements, duplicate entries, and incorrect sensor readings. Missing values were imputed with average values. Figure~\ref{fig:AvailableTimeSeries} shows the available minutely sampled time-series for ambient temperature, oil temperature, and load with a total of 5760 samples over 4 days of operation. Oil and winding temperature profiles analysed in this study are based on this minute-level sampling.

Since the PDE solution is unknown, no validation dataset is available for hyperparameter tuning. This makes balancing the loss function particularly challenging~\cite{wang2023expertsguidetrainingphysicsinformed}. To assess learning and generalization capabilities, the model is evaluated under varying numbers of initial condition samples $N_i$ (known values) and residual points $N_r$ (PDE evaluation points, unknown values), as detailed in Section~\ref{ss:Experiments}.

\subsection{Evaluation Metrics}
\label{ss:Evaluation Metrics}

The accuracy of probabilistic predictions is evaluated using three complementary metrics.

\vspace{-1 mm}

\noindent \textbf{Negative Log Likelihood} (NLL) evaluates how well a probabilistic model explains a given set of observations. It is defined as the negative logarithm of the likelihood function, which measures the probability of the observed data under a particular model~\cite{murphy_2012}. For a model with parameters $\bm{\theta}$ and observed data $X\! =\! \{x_1,\! \dots\!,\! x_n\!\}$, NLL is defined as:

\small

\vspace{-1 mm}

\begin{equation}
	\text{NLL}(\theta) = -\sum_{i=1}^n \log p(x_i \mid \bm{\theta})
	\label{eq:nll}
\end{equation}

\vspace{-1 mm}

\normalsize

\noindent where $p(x_i \mid \bm{\theta})$ is the probability of observation $x_i$ given the model parameters. Minimizing the NLL is equivalent to maximizing the likelihood function, as the logarithm is a monotonically increasing function. NLL is particularly useful because it avoids numerical underflow with small probabilities and transforms the product of probabilities into a sum of log probabilities, which is more computationally stable~\cite{bishop_2006}.

\noindent \textbf{Continuous Ranked Probability Score} (CRPS) measures the discrepancy between the predicted Cumulative Distribution Function (CDF), $F(\cdot)$, and the observed empirical CDF for a given scalar observation $y$~\cite{zamo2018}:

\small
\vspace{-1 mm}

\begin{equation}
	\label{eq:crps}
	CRPS(F,y) = \int (F(x) -\mathds{1}(x\geq y_i))^2 dx,
\end{equation}

\vspace{-1 mm}
\normalsize

\noindent where $\mathds{1}(x\geq y_i)$ is the indicator function, which models the empirical  CDF. 

In order to obtain a single score value from Eq.~(\ref{eq:crps}), a weighted average is computed for each individual observation of the test set~\cite{Gneiting2005}:

\small
\vspace{-1 mm}

\begin{equation}
	\label{eq:crps_avg}
	CRPS = \frac{1}{N} \sum_{i=1}^{N} CRPS(F_i,y_i)
\end{equation}
\vspace{-1 mm}
\normalsize

\noindent where $N$ denotes the total number of predictions.

\vspace{-1 mm}

\noindent \textbf{Prediction Interval Coverage Probability} (PICP) quantifies the reliability of prediction intervals by evaluating the fraction of true observations that fall within the predicted interval. PICP evaluates the effectiveness of the prediction interval in capturing the variability of the data, reflecting the accuracy and validity of the predictions~\cite{GONZALEZ2021}:

\small
\vspace{-1 mm}

\begin{equation}
	\label{eq:picp}
	PICP = \frac{1}{N} \sum_{i=1}^{N} \hat{\Theta}_{O_i}
\end{equation}
\vspace{-1 mm}
\normalsize

\noindent where $N$ denotes the total number of predictions and $\hat{\Theta}_{O_i}$ individual observations in the prediction interval.

\subsection{Benchmarking}

The proposed B-PINN approach is benchmarked against a probabilistic hybrid method. Namely, dropout based PINN implementation has been selected due to its computational efficiency \cite{Zhang_19}.

\subsection*{Dropout-PINN}


Dropout-PINN extends the vanilla PINN (Section~\ref{ss:PINN_Basics}) by inserting dropout layers after each hidden layer, with dropout rate $\rho$. During training, neurons are randomly deactivated, improving generalization and reducing overfitting~\cite{Gal16}. During inference, $K$ stochastic forward passes are performed, each with different neuron subsets deactivated. The resulting ensemble of predictions captures predictive uncertainty~\cite{Zhang_19}.

The dropout rate strongly affects the uncertainty quality \cite{Alcibar_2025} and therefore a sensitivity analysis is conducted to select the best dropout rate.

\subsection{Experiments and Implementation Details}
\label{ss:Experiments}

The base-case B-PINN uses two layers with 50 neurons each, $N_i$=100 initial points, $N_{bc}$=11520 boundary condition points, and $N_{r}$=10000 residual points. Input data noise and residual data noise follow $\sigma_i$=$\mathcal{N}(0,0.01)$ and $\sigma_r$=$\mathcal{N}(0,0.01)$, respectively.

Starting from the base-case, different hyperparameters are evaluated including the number of neurons and the number of initial and residual points. After identifying the best-performing configuration, experimental results are obtained for the selected configuration. In addition to the hyperparameters, three different priors are tested: Gaussian, Spike-and-Slab (cf. Eq.~(\ref{eq:spike_slab_prior})), and Laplace (cf. Eq.~(\ref{eq:L_prior})). Table~\ref{table:hyperpar} summarizes the configurations and corresponding hyperparameters.

\begin{table}[!htb]
	\centering
	\caption{Base-case and tested hyperparameter ranges for B-PINN models.}
	\begin{tabular}{|c|c|c|}
		\hline
		\multirow{2}{*}{\textbf{Parameters}}& \multicolumn{2}{|c|}{\textbf{Configurations}} \\ \cline{2-3}
		& \textbf{Base-Case} & \textbf{Range} \\ \hline
		Neurons & 50 & [10,25,50,100] \\ \cline{1-1}
		$N_i$ & 100 & [5,100,200] \\ \cline{1-1}
		$N_r$ & 10000 & [5000,10000, 20000] \\ \cline{1-1}
		$\sigma_i$ & 0.01 & [0.01, 0.05, 0.1] \\\cline{1-1}
		$\sigma_r$ & 0.01 & [0.01, 0.05, 0.1] \\\hline
	\end{tabular}
	\label{table:hyperpar}
\end{table}

All experiments use variational inference to approximate the posterior distribution. Each simulation is repeated five times for numerical stability, and mean and standard deviation values are reported. All models are implemented from scratch in PyTorch. The optimization algorithm is Adam (learning rate = 0.01, batch size = 16) and all B-PINN models are trained with 15000 epochs. The weights of the individual loss terms have been manually weighted, with final values of $\lambda_0$=1, $\lambda_b$=1, and $\lambda_r$=10\textsuperscript{-6}.

The vanilla PINN model is implemented using the Adam optimization algorithm for the first 20000 iterations, followed by the L-BFGS optimizer for the last 10000 iterations. The terms of individual loss terms have been manually weighted with final values of $\lambda_0$=1, $\lambda_b$=1, and $\lambda_r$=10\textsuperscript{-6}. Finally, the dropout-PINN model is designed taking the vanilla PINN architecture as the reference model, including additional dropout layers. The dropout rate is tested for a range of values $\rho$=[0.05, 0.1, 0.15, 0.2, 0.25], with $K$=200 Monte Carlo iterations.

\section{Results \& Discussion}
\label{sec:ResultsDiscussion}

\begin{figure}[!htb]
	\vspace{-0.2cm}
	\centering
	\includegraphics[width=.5\columnwidth]{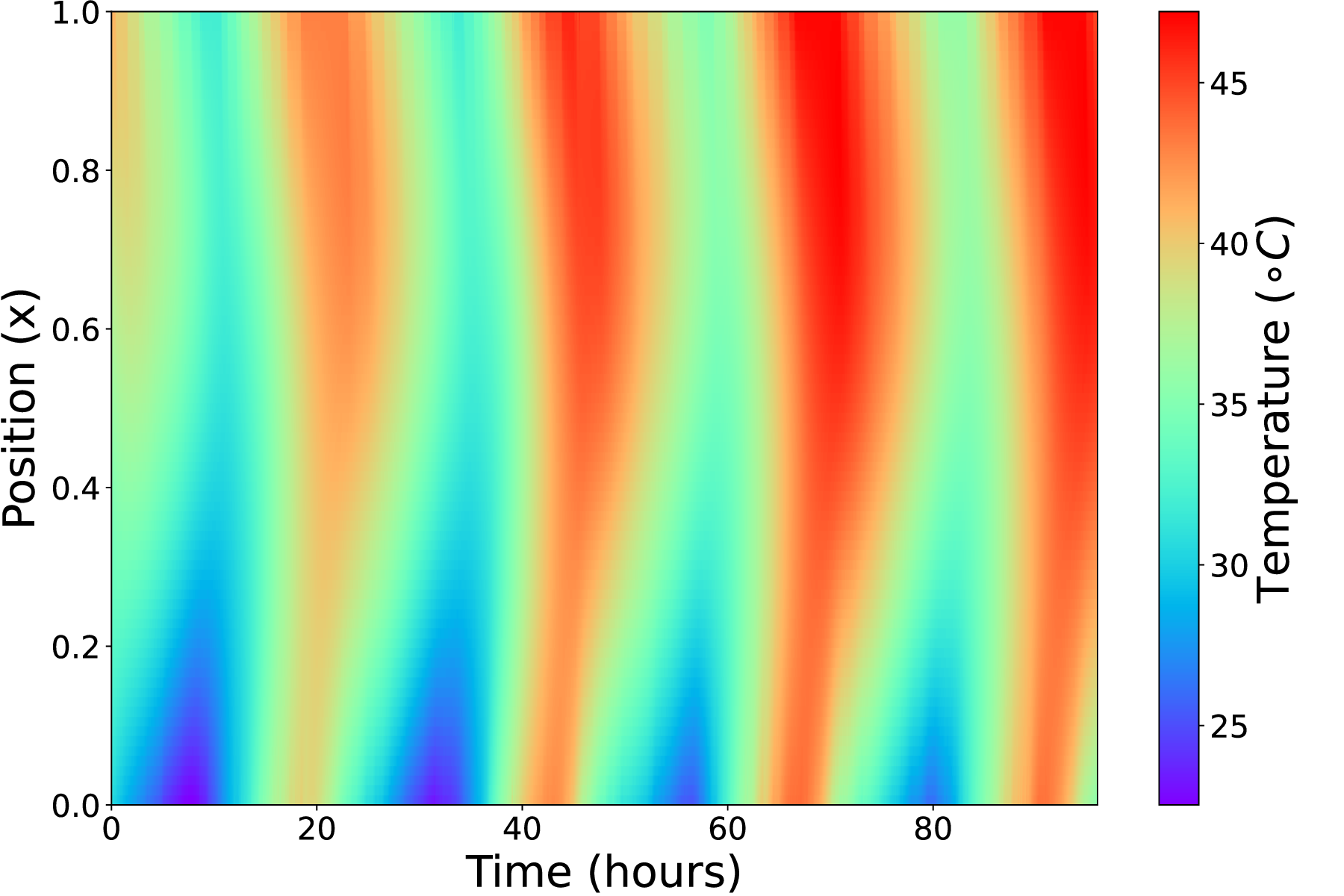}
	\caption{Spatiotemporal transformer oil temperature estimation obtained through a finite element method.}
	\label{fig:PDEmatlab}
	\vspace{-0.3cm}
\end{figure}

Figure~\ref{fig:PDEmatlab} shows the numerical solution of Eqs.~(\ref{eq:PDE_1D_Diffusion})--(\ref{eq:BoundaryConditions}), solved using finite element method (FEM) with Matlab's \texttt{pdepe} solver \cite{matlabpdepe}. The solution is experimentally validated against fiber optic sensor measurements installed at different transformer heights. Refer to \cite{Ramirez_25} for more details.

\begin{table*}[!htb]
	\centering
	\begin{threeparttable}
		\centering
		\small
		\caption{Comparison of different number of neurons and prior distributions for different UQ metrics with remaining parameters fixed  (2 layers, $N_i$=100, $N_r$=10000,  $\sigma_i$=$\sigma_r$=$\mathcal{N}(0,0.01^2)$). Best results highlighted.}
		\setlength{\tabcolsep}{2.000pt}
		\begin{tabular}{|c|ccc|ccc|ccc|}
			\hline
			\multirow{2}{*}{\textbf{\#N}} & \multicolumn{3}{c|}{\textbf{Gaussian}} & \multicolumn{3}{c|}{\textbf{Spike-and-Slab}}& \multicolumn{3}{c|}{\textbf{Laplace}} \\
			\cline{2-10}
			& PICP ($\uparrow$) & CRPS ($\downarrow$) & NLL($\downarrow$) & PICP ($\uparrow$) & CRPS($\downarrow$) & NLL ($\downarrow$) & PICP ($\uparrow$) & CRPS ($\downarrow$) & NLL ($\downarrow$) \\
			\hline
			10 & 0.078\footnotesize$\pm$.0225 & 0.287\footnotesize$\pm$.051 & 5.774\footnotesize$\pm$2.850 & 0.067\footnotesize$\pm$.012 & 0.365\footnotesize$\pm$.096 & 15.898\footnotesize$\pm$11.827 & 0.071\footnotesize$\pm$.018 & 0.297\footnotesize$\pm$.075 & 11.933\footnotesize$\pm$5.840 \\
			25 & 0.197\footnotesize$\pm$.028 & 0.180\footnotesize$\pm$.045 & 0.445\footnotesize$\pm$.608 & 0.159\footnotesize$\pm$.043 & 0.130\footnotesize$\pm$.021 & -0.063\footnotesize$\pm$0.136 & 0.170\footnotesize$\pm$.016 & 0.153\footnotesize$\pm$.034 & 0.136\footnotesize$\pm$.345 \\
			50 & 0.377\footnotesize$\pm$.034 & 0.159\footnotesize$\pm$.042 & 0.194\footnotesize$\pm$.323 & 0.351\footnotesize$\pm$.036 & 0.139\footnotesize$\pm$.054 & -0.023\footnotesize$\pm$.336 & 0.350\footnotesize$\pm$.032 & \cellcolor{selected}0.115\footnotesize$\pm$.024 & \cellcolor{selected}0.017\footnotesize$\pm$.353 \\
			100 & 0.465\footnotesize$\pm$.018 & 0.238\footnotesize$\pm$.125 & 0.927\footnotesize$\pm$1.066 & 0.495\footnotesize$\pm$.058 & 0.278\footnotesize$\pm$.205 & 0.654\footnotesize$\pm$.776 & \cellcolor{selected}0.463\footnotesize$\pm$.083 & 0.204\footnotesize$\pm$.048 & 0.402\footnotesize$\pm$.255 \\
			\hline
		\end{tabular}
		\begin{tablenotes}[flushleft]
			\item \textbf{Legend}. \#N: Number of neurons.
		\end{tablenotes}
		\label{table:UQ_neurons}
	\end{threeparttable}
\end{table*}



The oil temperature values, $\hat\Theta_{O}(x,t)$, follow the seasonality of solar energy, \textit{i.e.} peak values coincide with periods of maximum solar irradiation and vice-versa (cf. Figure~\ref{fig:AvailableTimeSeries}). As for vertical temperature resolution, the highest temperatures are observed near the top position ($x\!\rightarrow\!1$). 

The validated FEM model in Figure~\ref{fig:PDEmatlab} serves as ground truth. Next, starting from the base-case configuration, a detailed hyperparameter tuning evaluation will be performed to select an appropriate B-PINN configuration.

\subsection{Hyperparameter tuning}

\textbf{Neurons and prior distributions.} The influence of the network width (number of neurons) and prior distribution is first examined with the remaining parameters fixed according to the base-case configuration (cf. Subsection \ref{ss:Experiments}). Table~\ref{table:UQ_neurons} reports the results.

Table~\ref{table:UQ_neurons} shows that the best performance is obtained with 50 neurons, with the Laplace prior distribution. This setting achieves the lowest CRPS (0.1156$\pm$0.0242) and NLL (0.0179$\pm$0.3536), indicating both accurate and sharp predictive distributions. At the same time, it maintains a high PICP (0.3501$\pm$0.0327), confirming that predictive intervals are well-calibrated and consistent with the observed data. These results suggest that the Laplace prior, in combination with a moderate network width, provides an adequate balance between predictive accuracy and reliable UQ.

Across all priors, increasing the number of neurons improves performance up to a point. Specifically, moving from 10 to 50 neurons leads to substantial improvements across all UQ metrics, producing sharper distributions, better likelihood scores, and higher coverage. However, further increasing the width to 100 neurons does not consistently improve results. While PICP occasionally increases, this often comes at the cost of higher CRPS and NLL values, indicating a deterioration in the overall quality of the predictive distributions.  

In summary, the results highlight the need for a trade-off when selecting the network width with respect to the prior distribution. In this case, the configuration with 50 neurons under the Laplace prior provides the most robust compromise, offering strong predictive performance together with well-calibrated uncertainty estimates.

\begin{table*}[!htb]
	\centering
	\begin{threeparttable}
		\small
		\caption{Comparison of different number of samples and prior distributions for different UQ metrics with remaining parameters fixed  (2 layers, 50 neurons, $\sigma_i=\sigma_r=\mathcal{N}(0,0.01^2)$). Best results highlighted.}
		\setlength{\tabcolsep}{1.000pt}
		\begin{tabular}{|c|c|ccc|ccc|ccc|}
			\hline
			\multicolumn{2}{|c|}{\textbf{\# Samples}} & \multicolumn{3}{c|}{\textbf{Gaussian}} & \multicolumn{3}{c|}{\textbf{Spike-and-Slab}} & \multicolumn{3}{c|}{\textbf{Laplace}}\\
			\hline
			$N_i$ & $N_r$ & \small PICP ($\uparrow$) & \small CRPS ($\downarrow$) & \small NLL($\downarrow$) & \small PICP ($\uparrow$) & \small CRPS($\downarrow$) & \small NLL ($\downarrow$) & \small PICP ($\uparrow$) & \small CRPS ($\downarrow$) & \small NLL ($\downarrow$) \\\hline
			\multirow{3}{*}{$I_1$} & $R_1$  & 0.320{\footnotesize$\pm$.027} & 0.208{\footnotesize$\pm$.052} & 0.451{\footnotesize$\pm$.255} & 0.334{\footnotesize$\pm$.060} & 0.233{\footnotesize$\pm$.083} & 0.641{\footnotesize$\pm$.569} & 0.390{\footnotesize$\pm$.059} & 0.178{\footnotesize $\pm$.030} & 0.338{\footnotesize$\pm$.127} \\\cline{2-2}
			& $R_2$ &  0.409 {\footnotesize $\pm$.052} & 0.202 {\footnotesize $\pm$.022} & 0.390 {\footnotesize $\pm$.083} & 0.363 {\footnotesize$\pm$.040} & 0.184{\footnotesize$\pm$.058}  & 0.261{\footnotesize$\pm$.237} & 0.450{\footnotesize$\pm$.018} & 0.177{\footnotesize$\pm$.045} & 0.331{\footnotesize$\pm$.178}\\\cline{2-2}
			& $R_3$  &  0.387{\footnotesize$\pm$.099} & 0.192{\footnotesize$\pm$.043} & 0.325{\footnotesize$\pm$.216} & 0.511{\footnotesize $\pm$.070} & 0.175{\footnotesize$\pm$.038} & 0.227{\footnotesize$\pm$.203} & \cellcolor{selected}0.541{\footnotesize$\pm$.043} & 0.169{\footnotesize $\pm$.022} & 0.208{\footnotesize$\pm$.150} \\	\hline
			\multirow{3}{*}{$I_2$} & $R_1$  &  0.282{\footnotesize $\pm$.044}& 0.157{\footnotesize $\pm$.035} & 0.173{\footnotesize $\pm$.183}& 0.322{\footnotesize $\pm$.042} & 0.187{\footnotesize$\pm$.048} & 0.358{\footnotesize $\pm$.226}& 0.305{\footnotesize$\pm$.032} & 0.113{\footnotesize$\pm$.027} & -0.096{\footnotesize$\pm$.208}\\\cline{2-2}
			& $R_2$  &  0.363{\footnotesize $\pm$.022}& 0.163{\footnotesize $\pm$.046} & 0.236{\footnotesize $\pm$.349} & 0.351{\footnotesize $\pm$.036}& 0.139{\footnotesize $\pm$.054} & -0.023{\footnotesize $\pm$.336} & 0.360{\footnotesize $\pm$.027} & 0.108{\footnotesize $\pm$.022} & -0.145{\footnotesize $\pm$.150} \\ \cline{2-2}
			& $R_3$ &  0.448{\footnotesize $\pm$.097} & 0.119{\footnotesize $\pm$.023} & -0.202{\footnotesize $\pm$.126} & 0.375{\footnotesize $\pm$.076}& 0.150 {\footnotesize $\pm$.050}& 0.080{\footnotesize $\pm$.346} & 0.377{\footnotesize $\pm$.040}  & 0.128{\footnotesize $\pm$.033} & -0.112{\footnotesize $\pm$.194} \\	\hline
			\multirow{3}{*}{\textbf{$I_3$}} & $R_1$  &  0.281{\footnotesize $\pm$.044} & 0.125{\footnotesize $\pm$.032} & -0.058{\footnotesize $\pm$.307} & 0.234{\footnotesize $\pm$.022} & 0.109{\footnotesize $\pm$.061} & -0.063{\footnotesize $\pm$.720} & 0.289{\footnotesize $\pm$.036} & 0.134{\footnotesize $\pm$.046} & -0.009{\footnotesize $\pm$.247} \\ \cline{2-2}
			& $R_2$&  0.416{\footnotesize $\pm$.041} & 0.092{\footnotesize $\pm$.026} & -0.374{\footnotesize$\pm$.218} & 0.348{\footnotesize $\pm$.083} & 0.131{\footnotesize $\pm$.045} & 0.127{\footnotesize $\pm$.733} &  0.324 {\footnotesize $\pm$.034}& \cellcolor{selected}0.077{\footnotesize $\pm$.009} & \cellcolor{selected}-0.478{\footnotesize$\pm$.071}\\ \cline{2-2}
			& $R_3$  &  0.435{\footnotesize $\pm$.072} & 0.166{\footnotesize $\pm$.051} & 0.455{\footnotesize $\pm$.590} & 0.390{\footnotesize $\pm$.082} & 0.121{\footnotesize $\pm$.048} & -0.113{\footnotesize $\pm$.414} & 0.420{\footnotesize $\pm$.042} & 0.137{\footnotesize $\pm$.034} & -0.050{\footnotesize $\pm$.336} \\	\hline
		\end{tabular}
		\begin{tablenotes}[flushleft]
			\item \textbf{Legend}. \# of initial samples ($N_i$): $I_1$=5, $I_2$=100, $I_3$=200. \# of residual samples ($N_r$): $R_1$=500, $R_2$=10000, $R_3$=20000.
		\end{tablenotes}
		\label{table:UQ_NumSamples}
	\end{threeparttable}
\end{table*}

\textbf{Number of initial ($N_i$), residual samples ($N_r$), and prior distributions.} Table~\ref{table:UQ_NumSamples} presents the sensitivity of the B-PINN model to the number of initial conditions samples, residual points, and boundary conditions, evaluated under different prior distributions using probabilistic metrics (cf. Subsection \ref{ss:Experiments}). 

The results indicate that increasing the number of initial condition samples generally improves the quality of uncertainty estimates across all priors. For instance, under the Gaussian prior, increasing $N_i$ from 5 to 200 (with fixed $N_r$=10000) reduces the NLL from 0.390$\pm$0.083 to -0.374$\pm$0.218, while also lowering the CRPS. This pattern is consistent across priors, suggesting that larger $N_i$ values leads to more informative predictive distribution, reflected in higher coverage (PICP) and improved probabilistic accuracy (CRPS, NLL).

In contrast, increasing the number residual points does not consistently enhance performance. For example, under the spike-and-slab prior with $N_i$=5, increasing $N_r$ from 10000 to 20000 results in only a marginal NLL improvement (0.261$\pm$0.237 to 0.227$\pm$0.203), with similar trends observed for other priors. These results suggest that excessively increasing the residual samples may reduce the relative influence of initial and boundary data, potentially degrading uncertainty calibration.

Under the adopted loss weighing scheme (cf. Eq.~(\ref{eq:ELBO_BPINN}), Algorithm~\ref{alg:Bayesian_PINN}), this trade-off highlights the need for a balanced allocation between residual points and initial/boundary condition samples. Residual points remain essential to enforce PDE dynamics, but their dominance over data-driven terms can compromise calibration. Hence, an appropriate trade-off between $N_r$ and $N_i$ is critical to achieve reliable predictive performance.

The best results are achieved with 200 initial samples and 10000 residual points, using the Laplace prior distribution. Consequently, subsequent experiments will be performed with the selected configuration.

\subsection{Experimental Results}

\begin{figure}[!htb]
	\centering
	\subfloat[]{\includegraphics[width=.45\columnwidth]{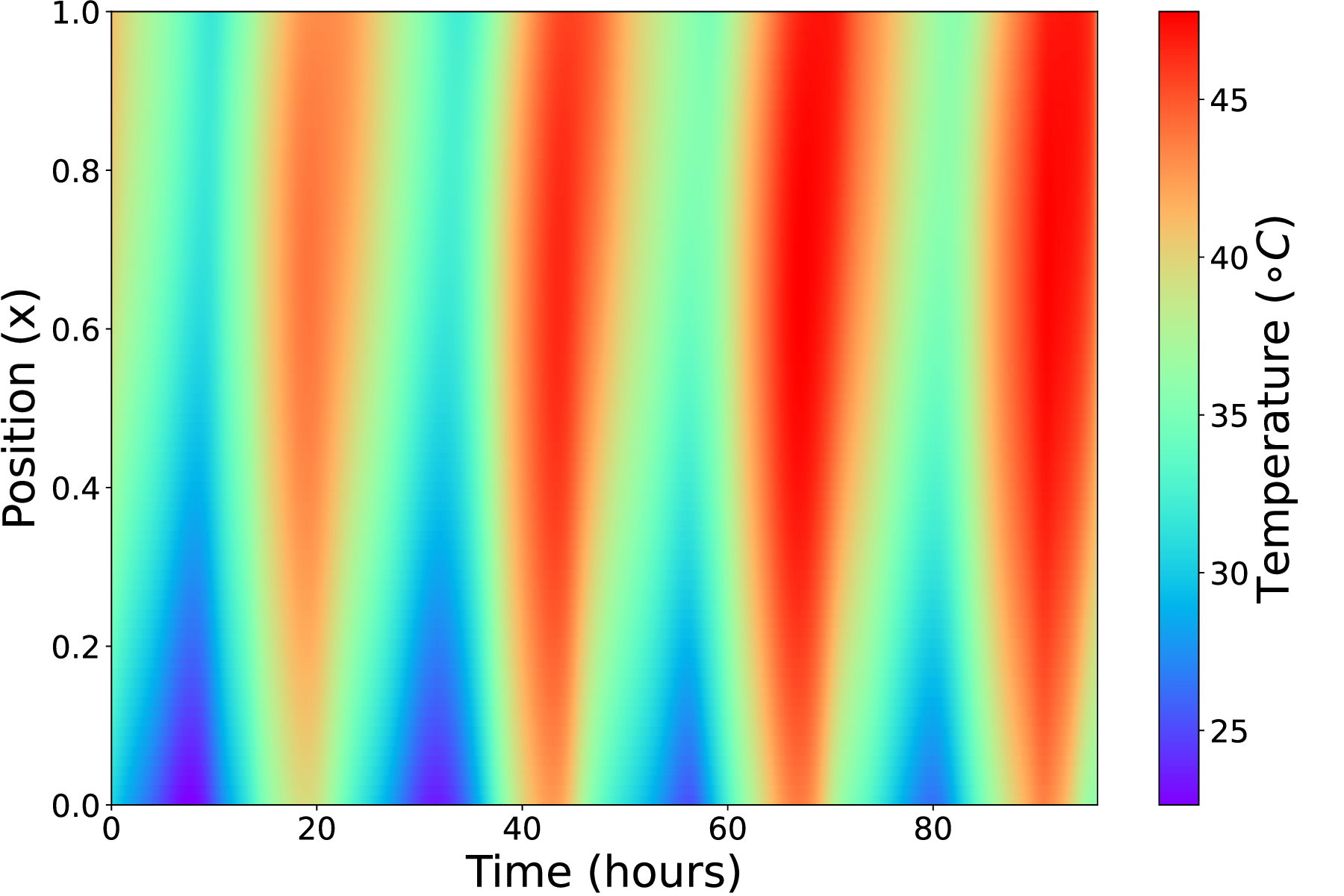}\label{subfig:temp_prediction_vanilla}} 
	\hspace{0.01\textwidth}
	\subfloat[]{\includegraphics[width=.45\columnwidth]{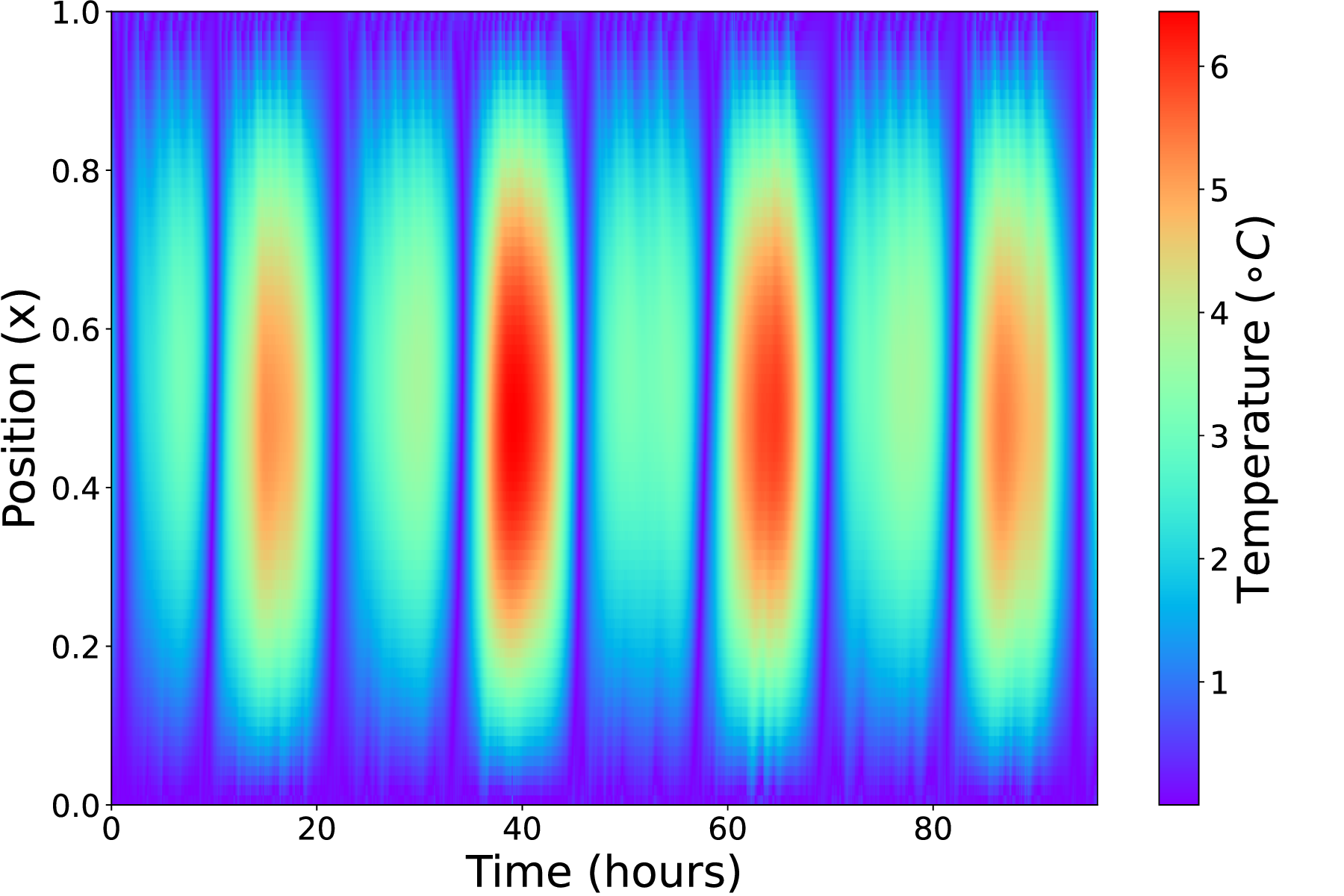}\label{subfig:temp_error_vanilla}}
	\caption{Spatiotemporal temperature estimation results obtained using a vanilla PINN. (a) Predicted temperature distribution across the transformer oil domain and (b) corresponding absolute error.}
	\label{fig:TemperatureEstimation_PINN}
\end{figure}

First, deterministic spatiotemporal transformer oil temperature estimates are obtained using the vanilla PINN (Section~\ref{ss:PINN_Basics}) following the implementation in \cite{Ramirez_25}. Figure~\ref{fig:TemperatureEstimation_PINN}(a) shows the spatiotemporal temperature estimate and Figure~\ref{fig:TemperatureEstimation_PINN}(b) shows the error with respect to the ground truth (Figure~\ref{fig:PDEmatlab}).

To compare the B-PINN outcome with a probabilistic estimate, a dropout-PINN model has been implemented following the PINN architecture (3 layers, 20 neurons each). Table~\ref{table:Dropout_PINN} shows the results for different dropout rates.

\begin{table}[!htb]
	\small
	\centering
	\caption{Sensitivity of the dropout rate in dropout-PINN. Best results highlighted.}
	\setlength{\tabcolsep}{5.000pt}
	\begin{tabular}{|c|ccc|} \hline Dropout Rate & PICP ($\uparrow$) & CRPS ($\downarrow$) & NLL($\downarrow$) \\ \hline 
		0.05 & 0.714 {\footnotesize $\pm$.045} &  \cellcolor{selected} 0.722 {\footnotesize $\pm$.044} & \cellcolor{selected} 1.955 {\footnotesize $\pm$.159} \\ 
		0.1 & 0.776 {\footnotesize $\pm$.061} & 0.853 {\footnotesize $\pm$.031} & 2.010 {\footnotesize $\pm$.143} \\ 
		0.15 & \cellcolor{selected}0.805 {\footnotesize $\pm$.071} & 0.967 {\footnotesize $\pm$.060} & 2.248 {\footnotesize $\pm$.265} \\ 
		0.2 & 0.787 {\footnotesize $\pm$.013} & 1.069 {\footnotesize $\pm$.036} & 2.386 {\footnotesize $\pm$.157} \\ 
		0.25 & 0.724 {\footnotesize $\pm$.048} & 1.214 {\footnotesize $\pm$.019} & 2.731 {\footnotesize $\pm$.226} \\ \hline 
	\end{tabular}
	\label{table:Dropout_PINN}
\end{table}

\begin{figure*}[!htb]
	\centering
	\subfloat[]{\includegraphics[width=0.45\columnwidth]{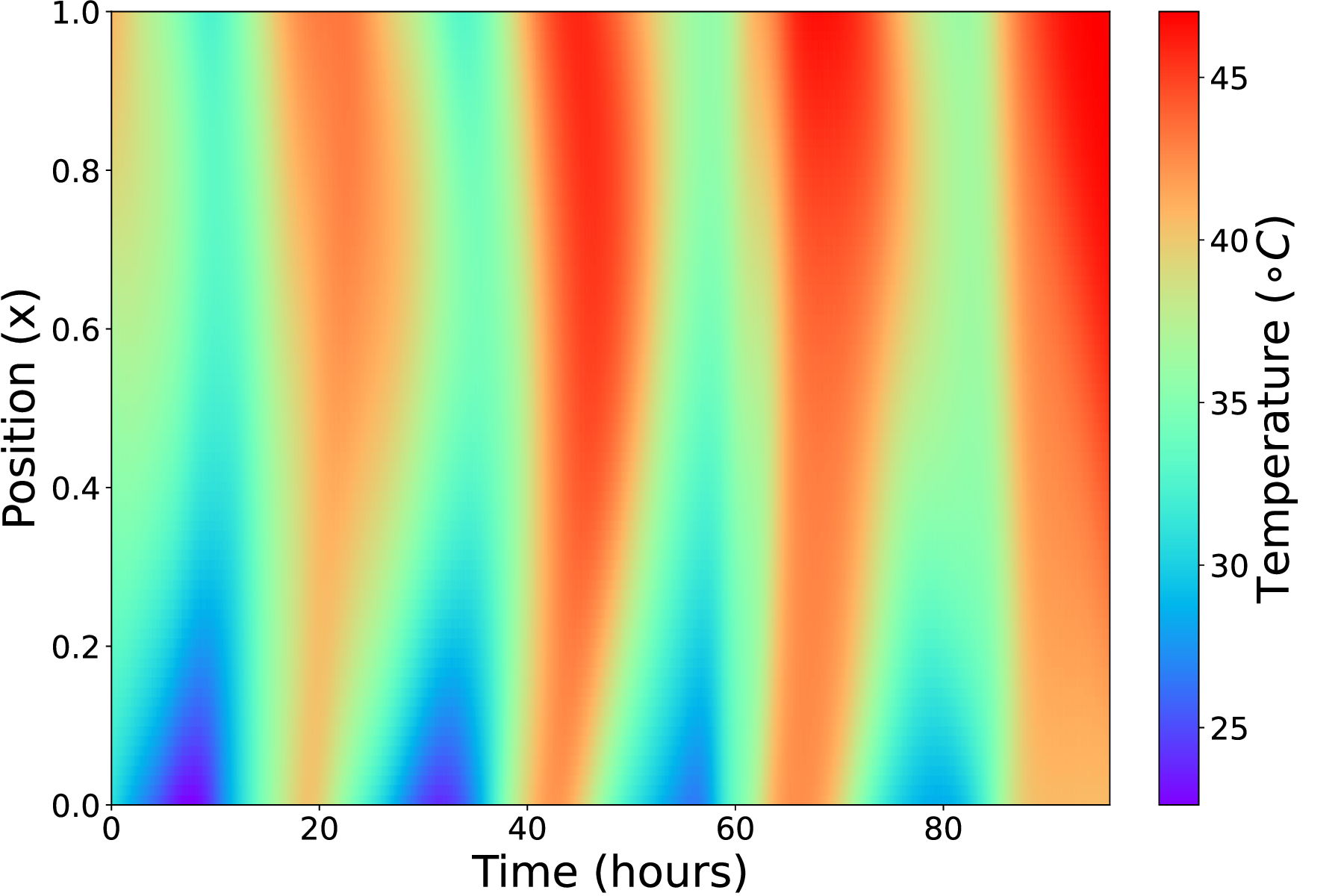}\label{subfig:temp_mean}}%
	\hspace{0.02\textwidth} 
	\subfloat[]{\includegraphics[width=0.45\columnwidth]{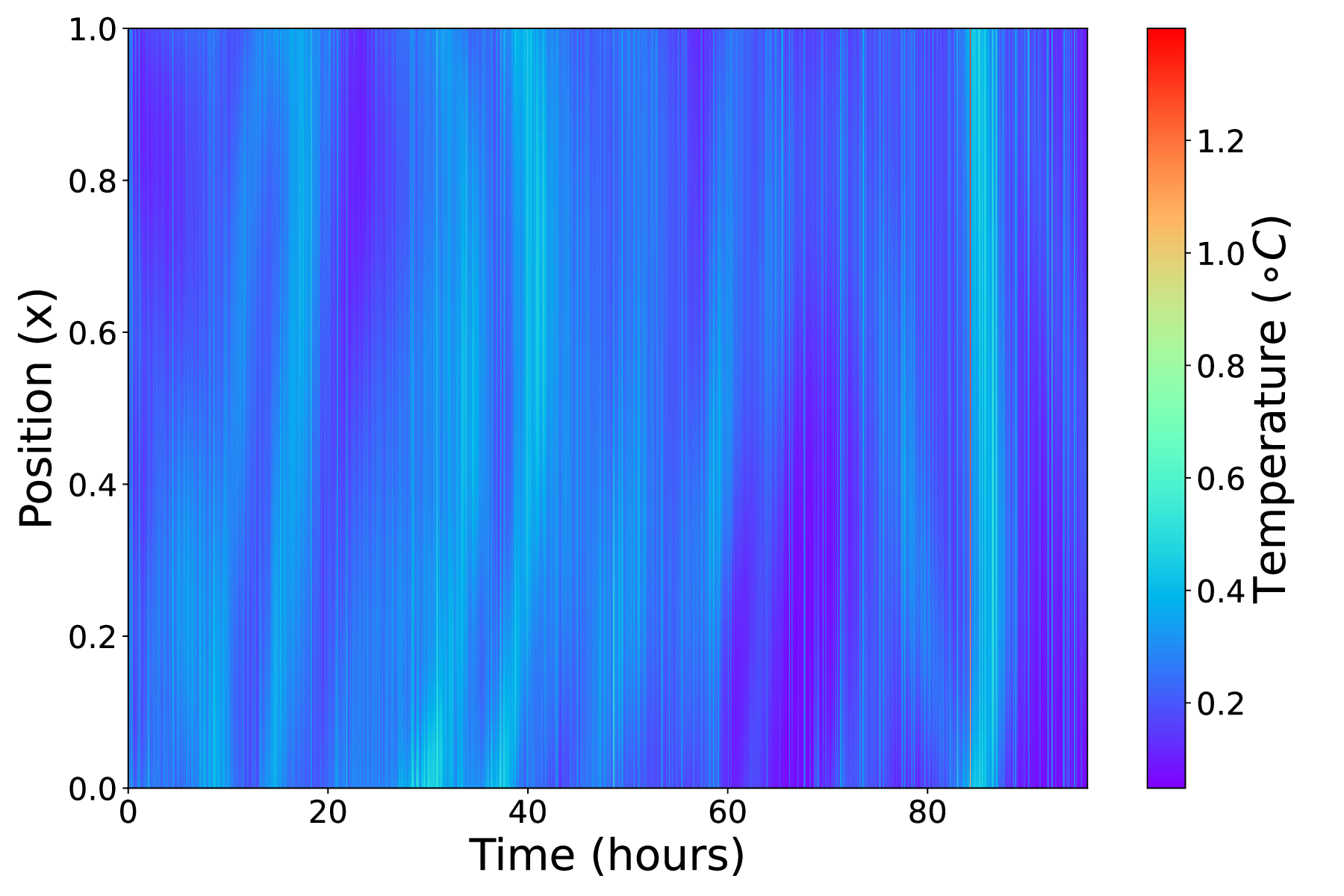}\label{subfig:temp_std}}\\[1ex] 
	\subfloat[]{\includegraphics[width=0.45\columnwidth]{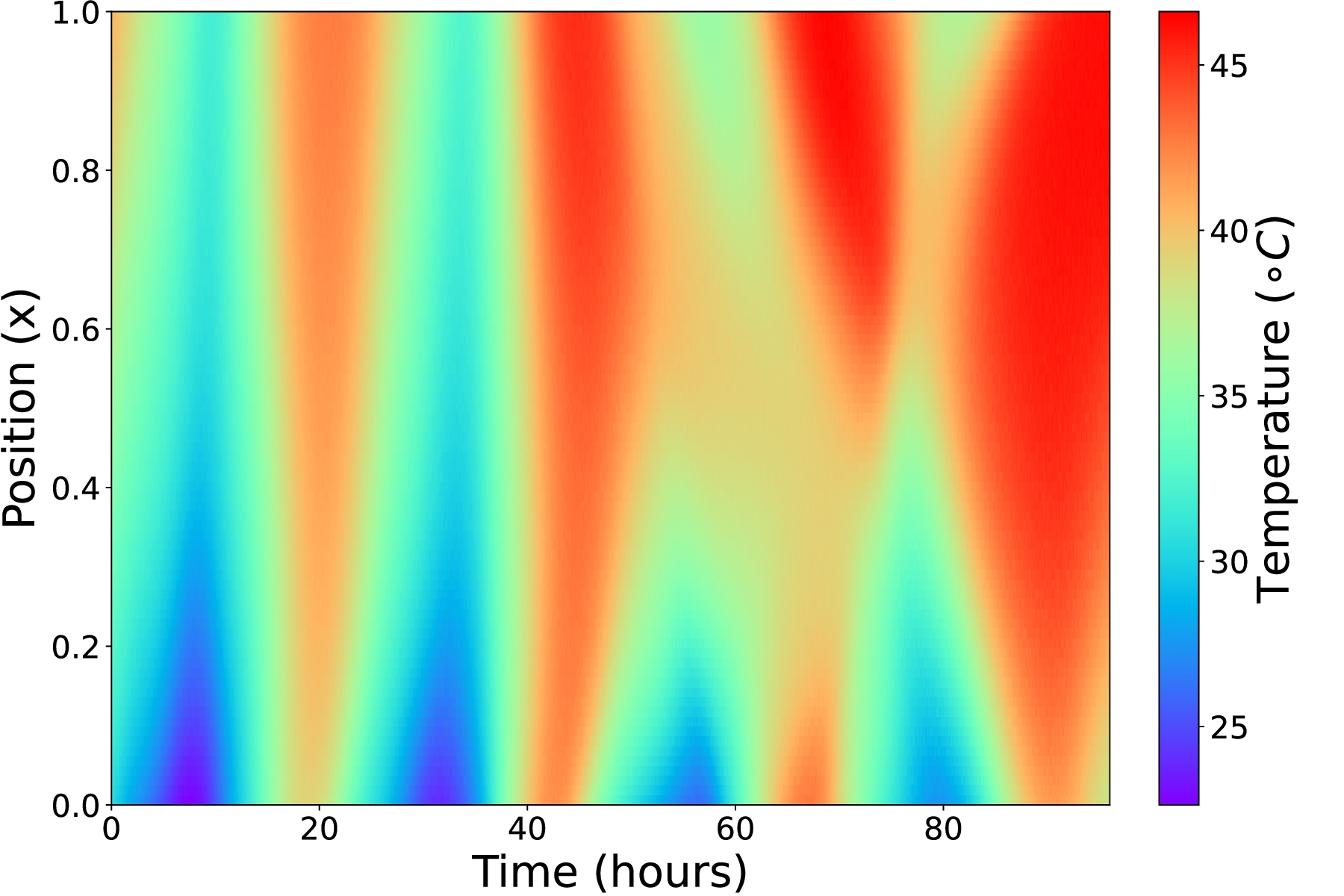}\label{subfig:temp_mean_MCD}}%
	\hspace{0.02\textwidth} 
	\subfloat[]{\includegraphics[width=0.45\columnwidth]{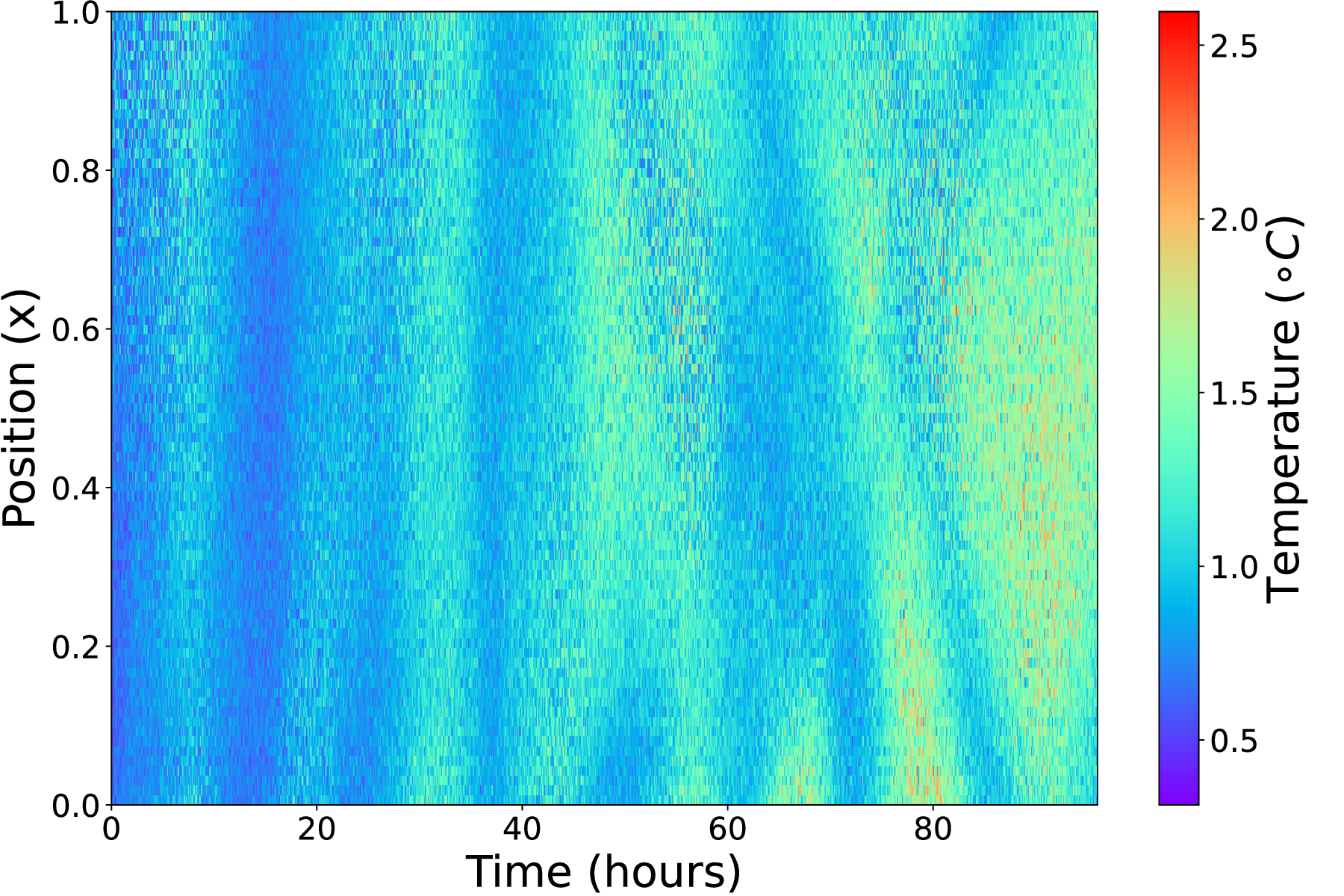}\label{subfig:temp_std_MCD}}
	\caption{Spatiotemporal transformer oil temperature estimation using the B-PINN model (a) posterior mean; (b) standard deviation; and dropout-PINN model (c) posterior mean, (d) posterior standard deviation.}
	\label{fig:MeanTemperatureEstimation_BPINN_MCD}
\end{figure*}

The best-performing dropout-PINN is obtained with a dropout rate of 0.05. Higher dropout rates achieve better PICP values, but at the cost of degraded CRPS and NLL performance. Accordingly, the best dropout-PINN and B-PINN models were selected for comparison. Figure~\ref{fig:MeanTemperatureEstimation_BPINN_MCD} shows the resulting probabilistic spatiotemporal temperature estimates parameterized through the mean and standard deviation.

\begin{figure*}[!htb]
	\centering
	\subfloat[]{\includegraphics[width=0.45\columnwidth]{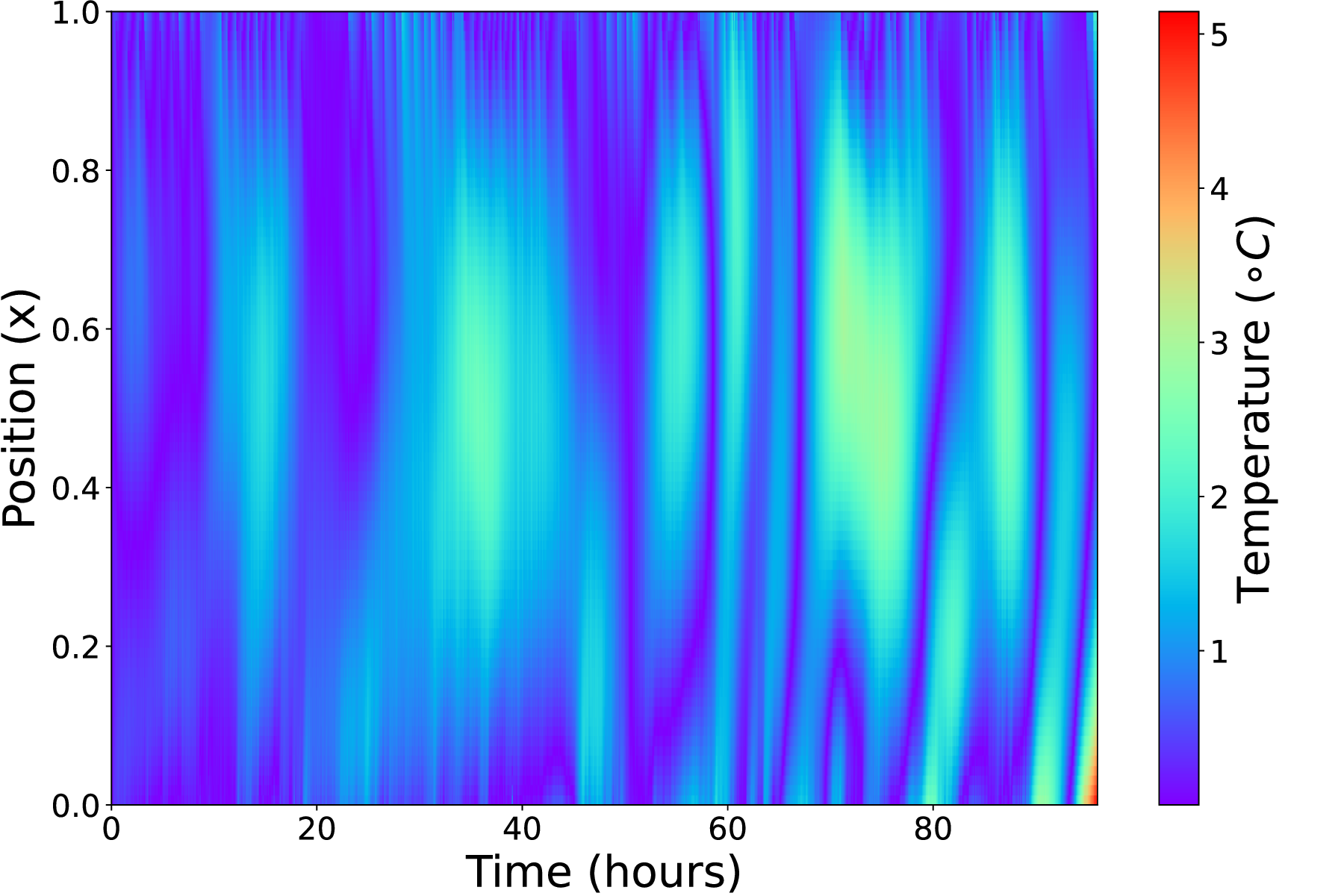}\label{subfig:temp_error_mean}}%
	\hspace{0.02\textwidth} 
	\subfloat[]{\includegraphics[width=0.45\columnwidth]{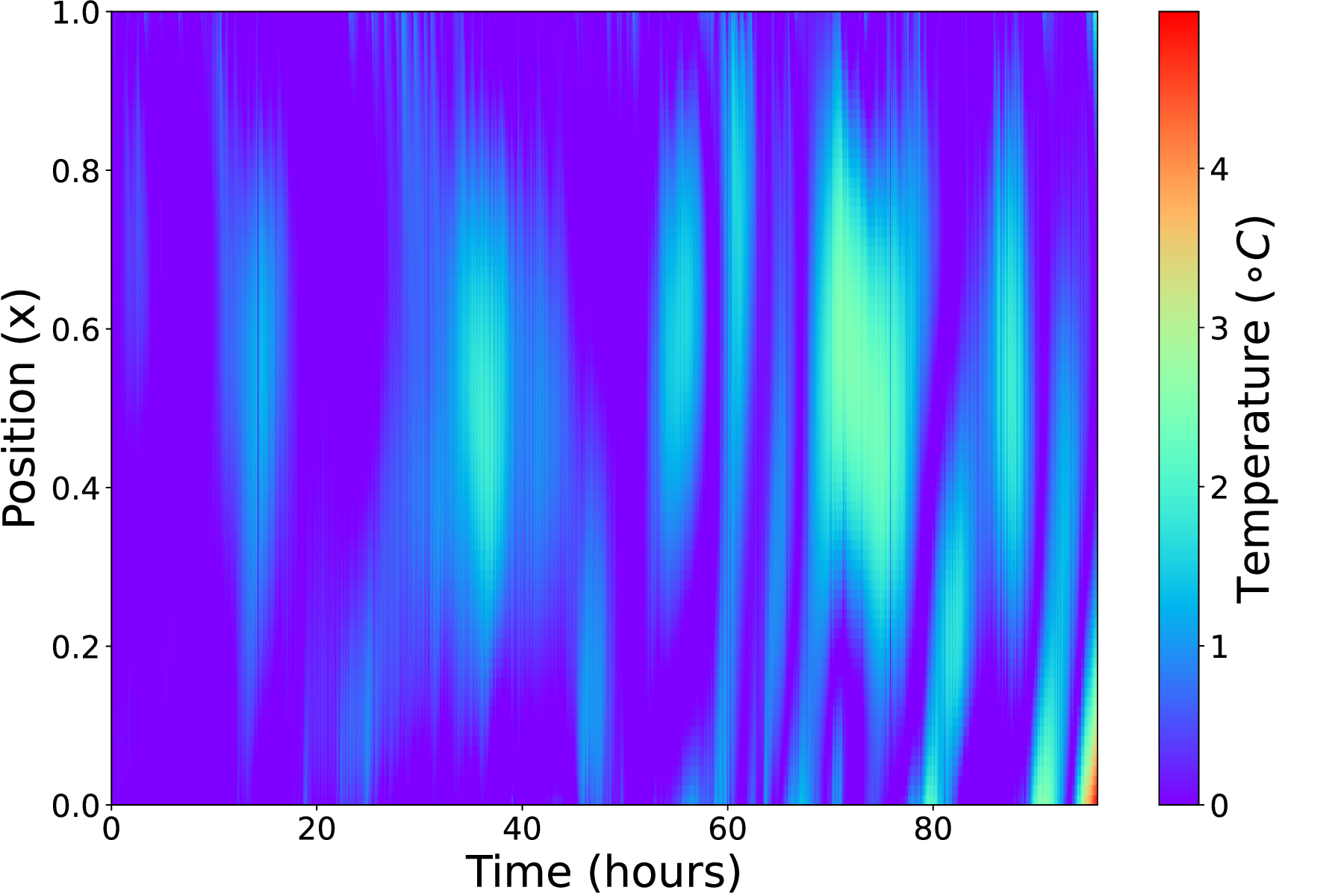}\label{subfig:temp_error_std_BPÌNN}}\\[1ex] 
	\subfloat[]{\includegraphics[width=0.45\columnwidth]{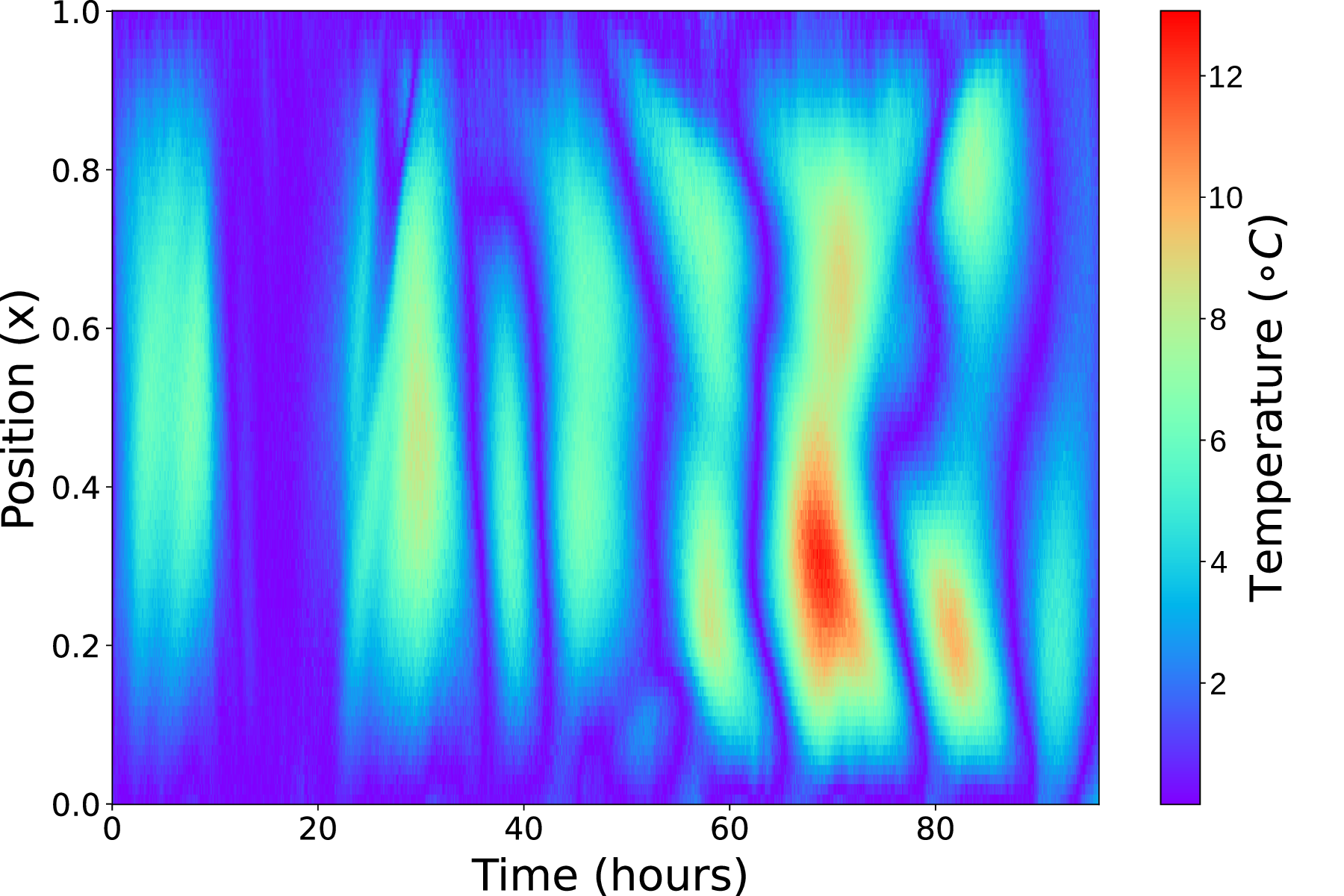}\label{subfig:temp_error_mean_MCD}}%
	\hspace{0.02\textwidth} 
	\subfloat[]{\includegraphics[width=0.45\columnwidth]{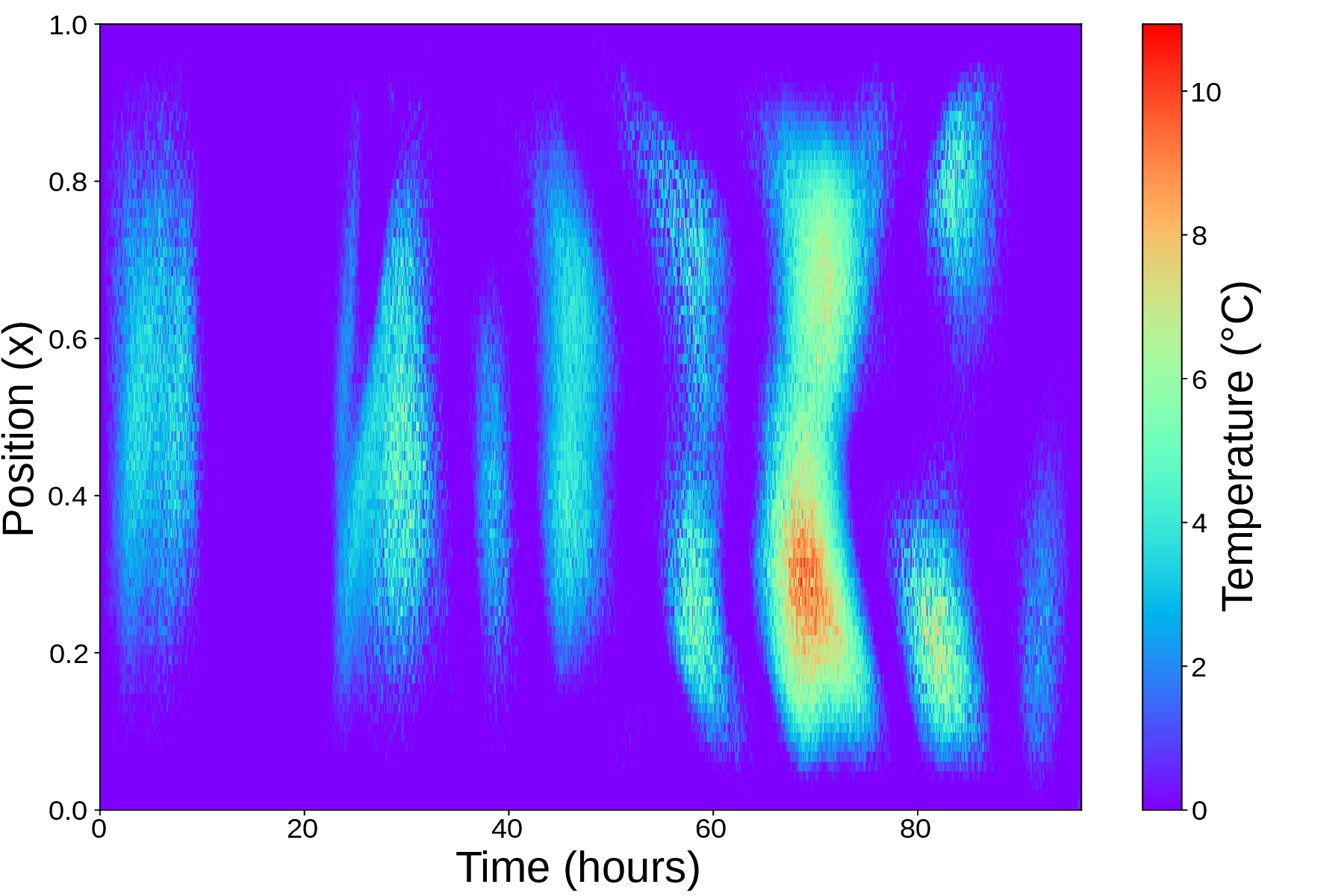}\label{subfig:temp_error_std_MCD}}
	\caption{Spatiotemporal transformer oil temperature estimation error of the B-PINN model (a) mean absolute error and (b) standard deviation; and dropout PINN model (c) mean absolute error and (d) standard deviation.}
	\label{fig:MeanTemperatureEstimationError_BPINN_MCD}
\end{figure*}

Compared with the vanilla PINN (Figure~\ref{fig:TemperatureEstimation_PINN}), the B-PINN is able to capture variation in the mean estimates, thereby quantifying prediction confidence at each spatiotemporal coordinate. The mean B-PINN predictions exhibit higher accuracy than the dropout-PINN, as also reflected in the error maps of Figure~\ref{fig:MeanTemperatureEstimationError_BPINN_MCD}(a) and (c). Furthermore, the variance of B-PINN predictions is consistently lower than dropout-PINN, confirmed in Figures~\ref{fig:MeanTemperatureEstimationError_BPINN_MCD}(b) and (d), which illustrate the standard deviation of the prediction error relative to the ground truth (cf. Figure~\ref{fig:PDEmatlab}).

\begin{figure*}[!htb]
	\centering
	\includegraphics[width=0.33\columnwidth]{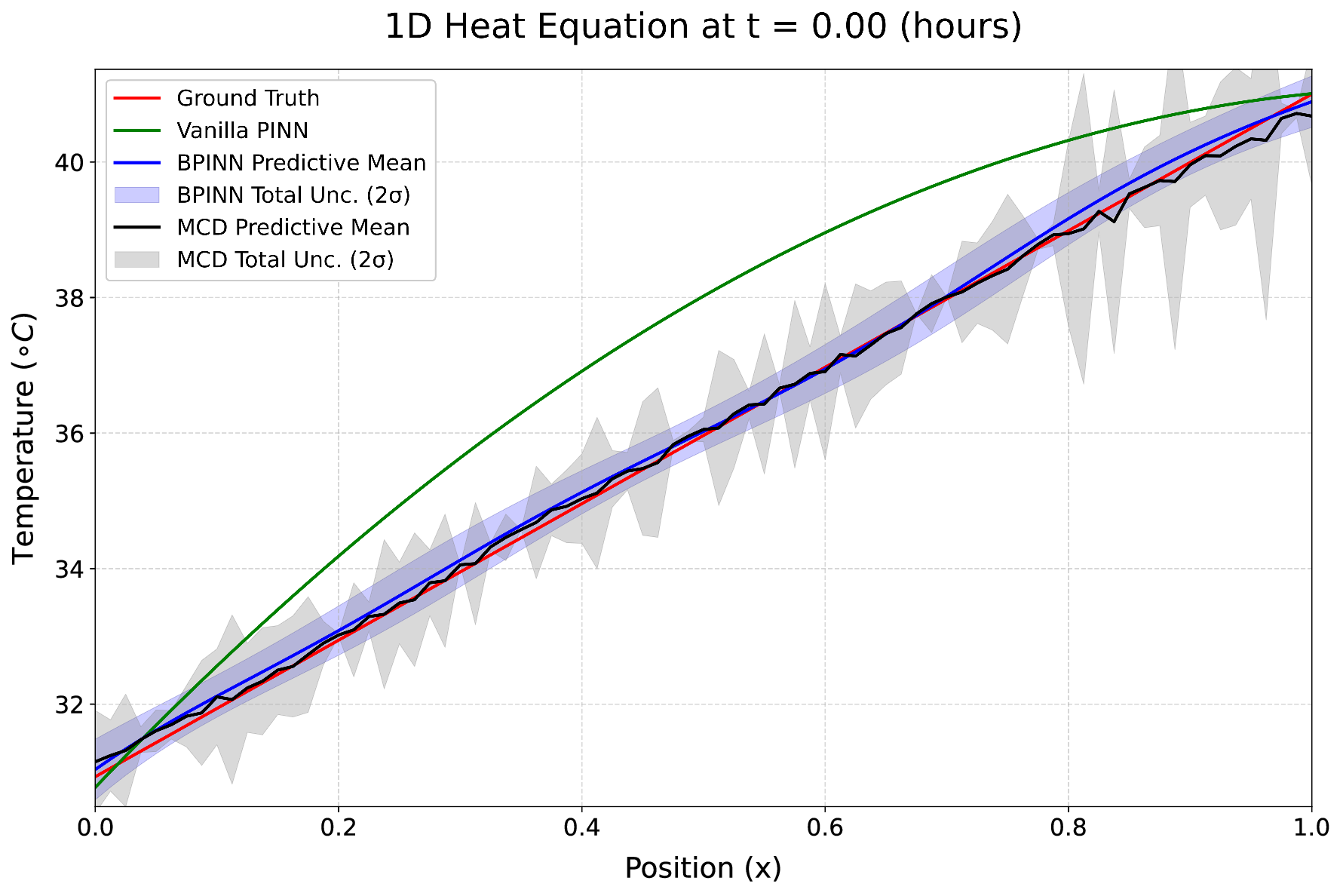}
	\includegraphics[width=0.33\columnwidth]{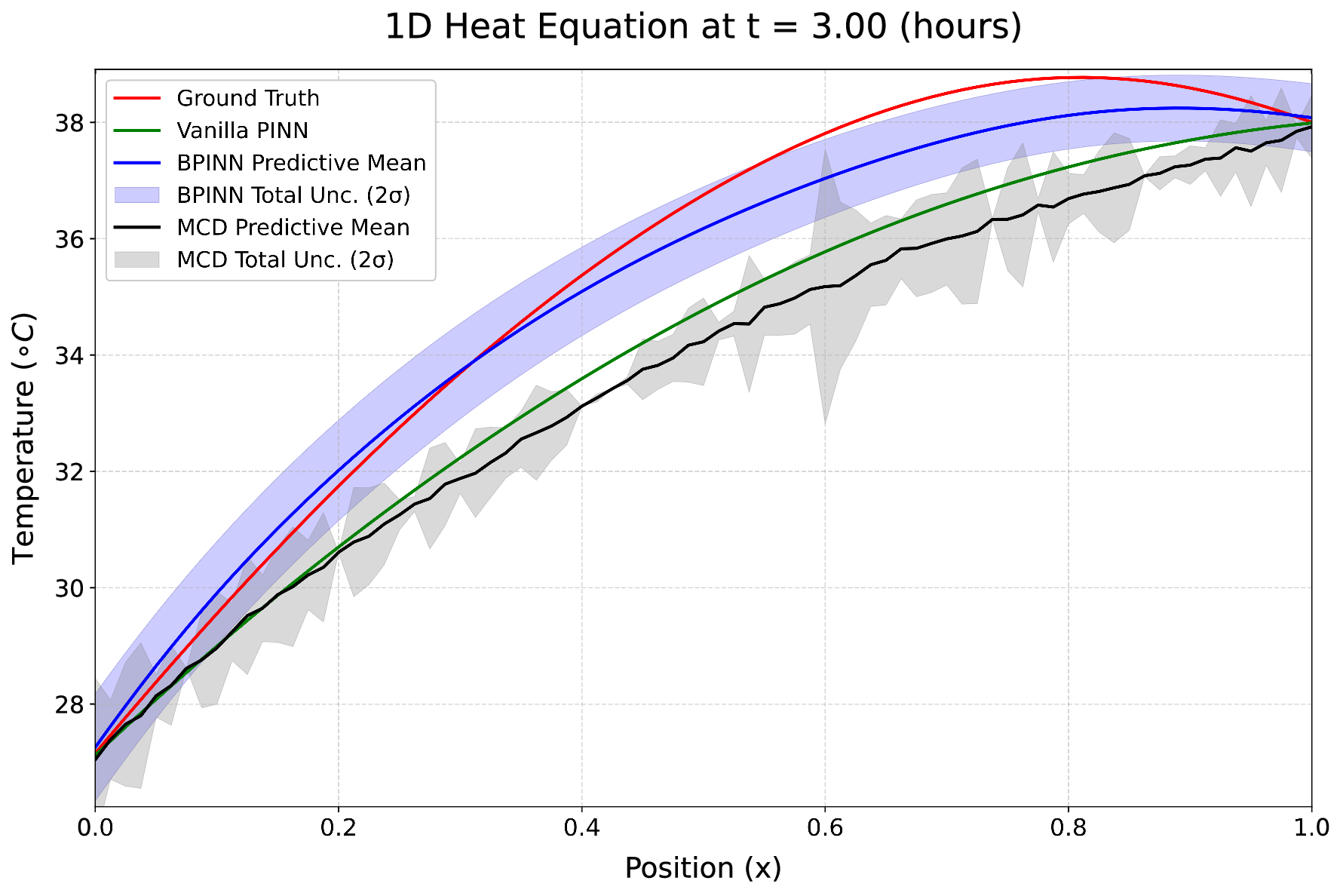}
	\includegraphics[width=0.33\columnwidth]{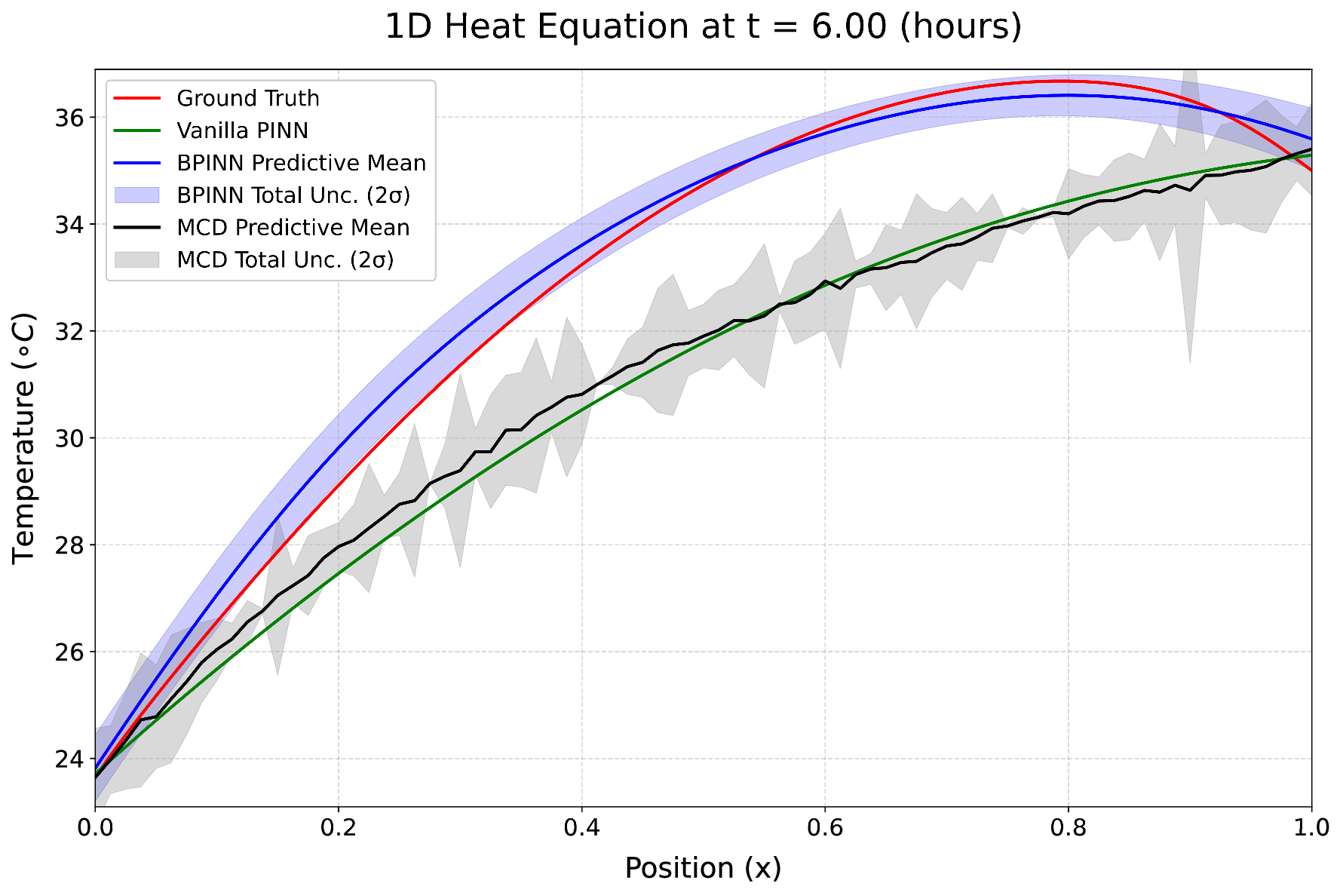}\\[0.5em]
	
	\includegraphics[width=0.33\columnwidth]{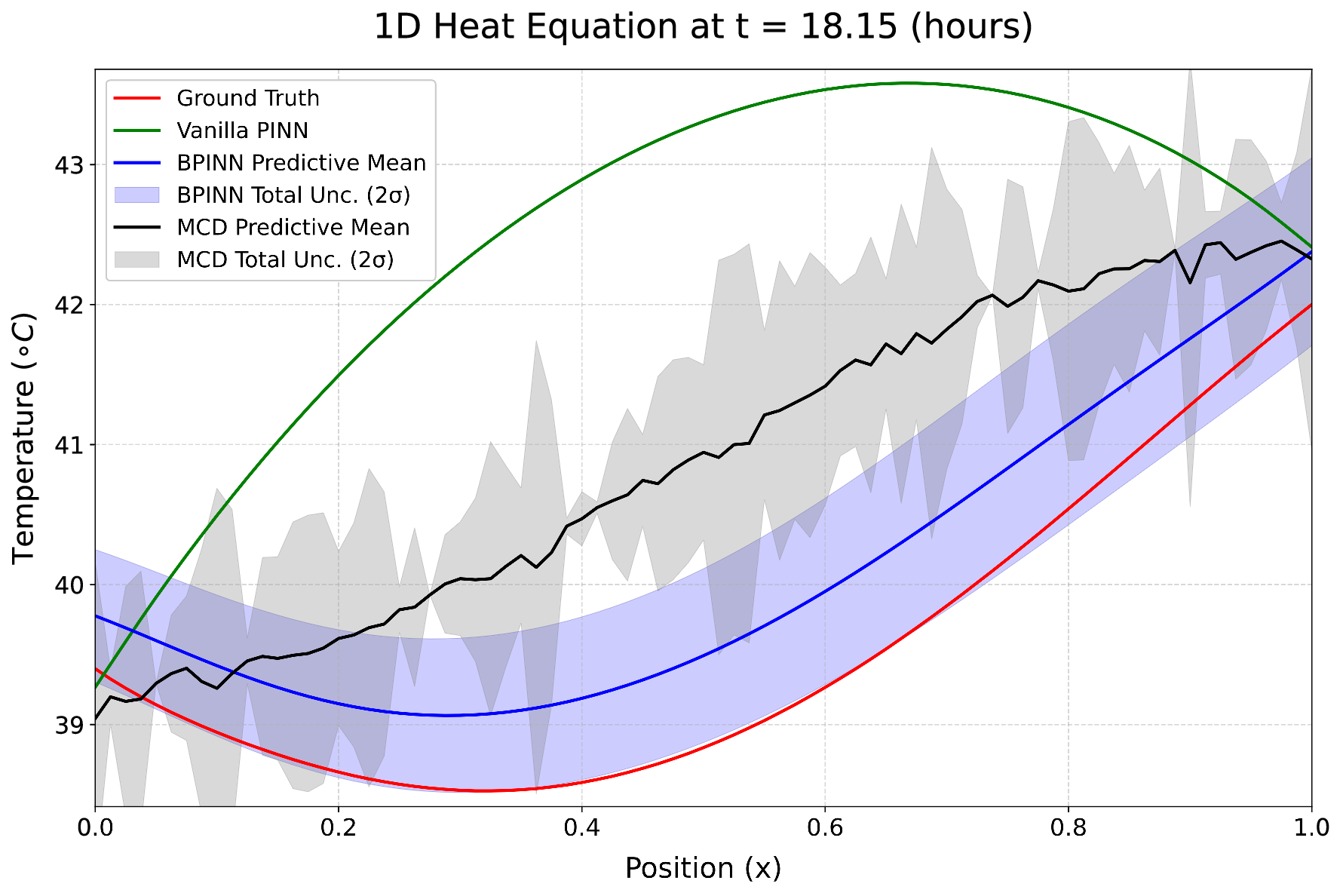}
	\includegraphics[width=0.33\columnwidth]{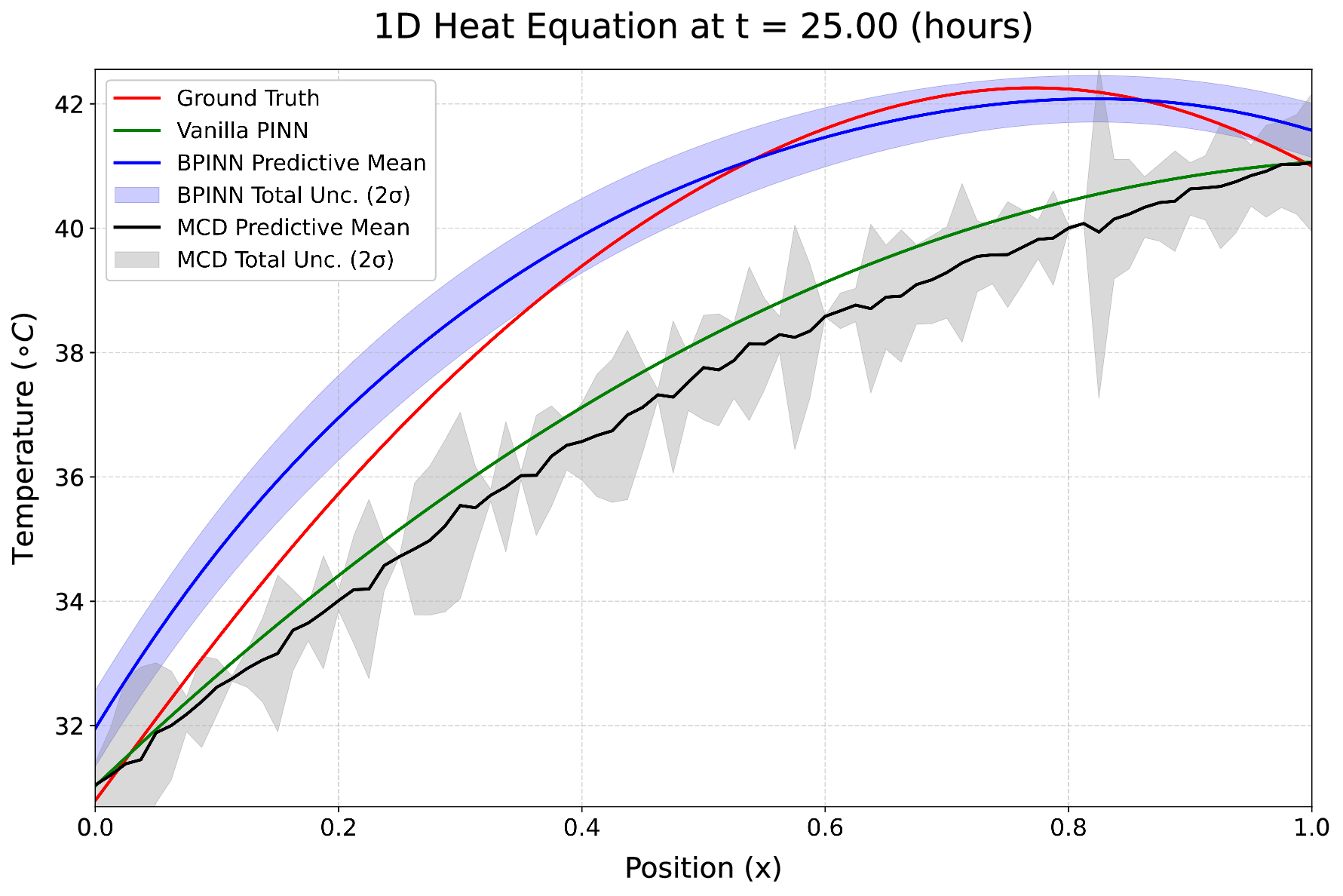}
	\includegraphics[width=0.33\columnwidth]{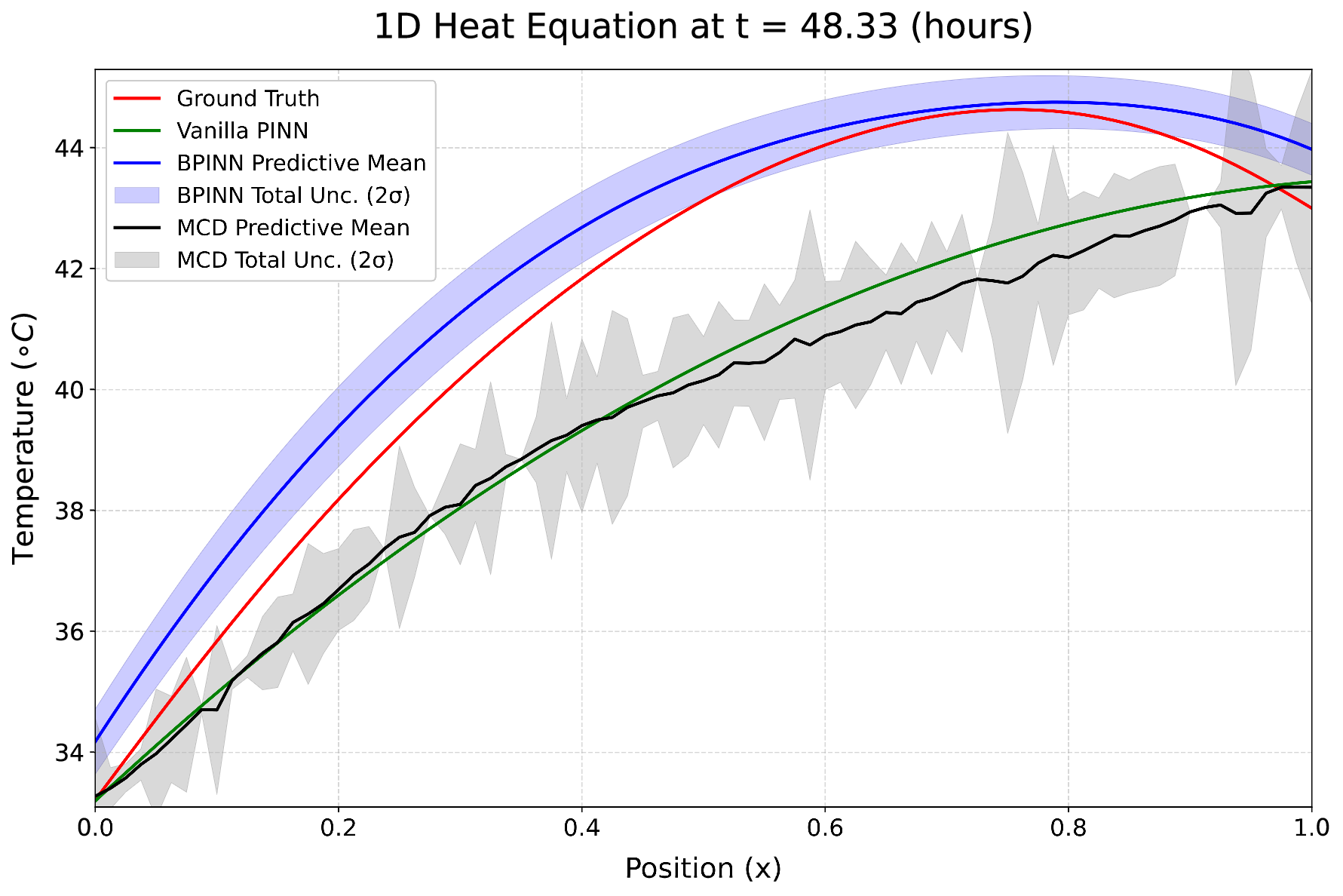}\\[0.5em]
	
	\includegraphics[width=0.33\columnwidth]{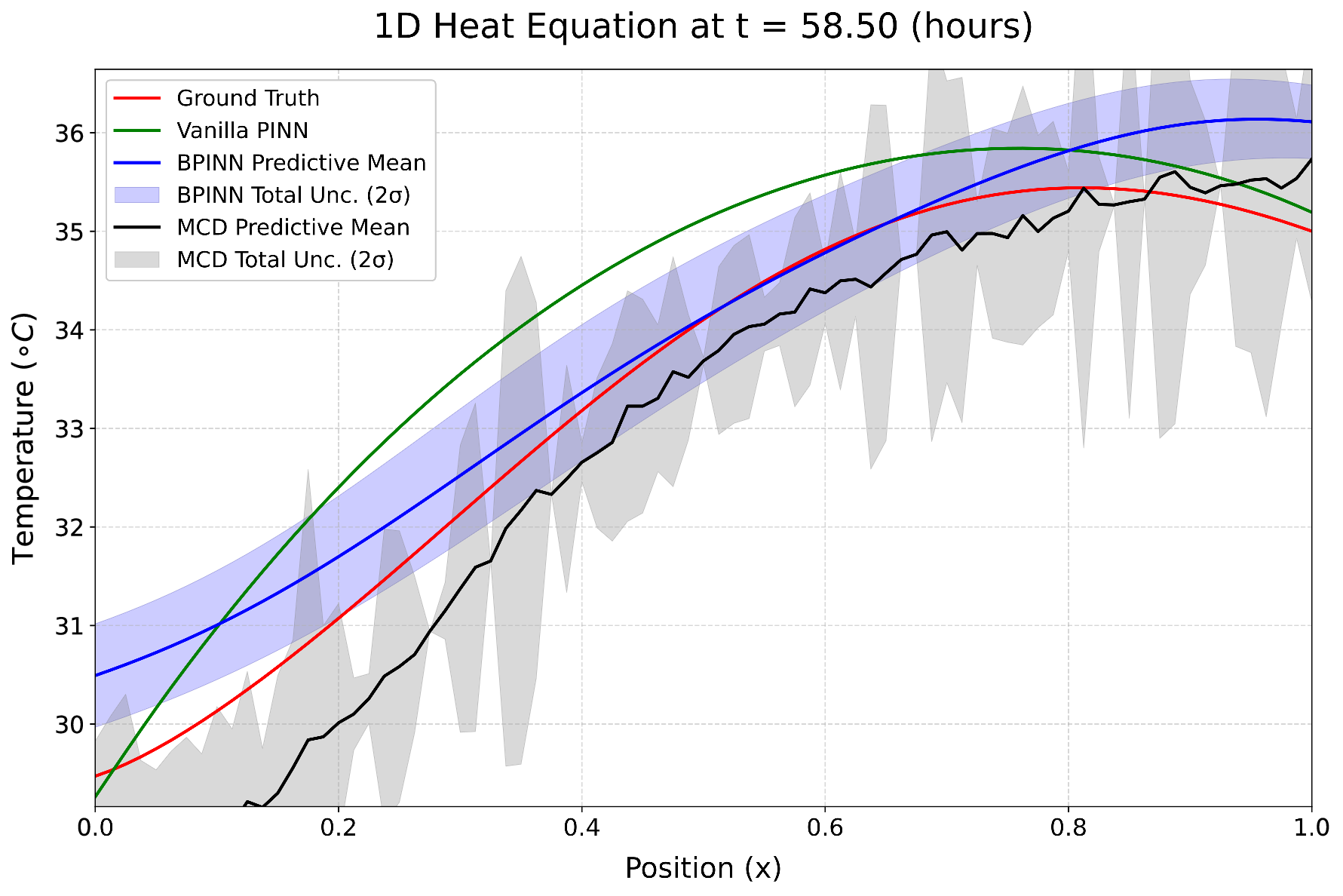}
	\includegraphics[width=0.33\columnwidth]{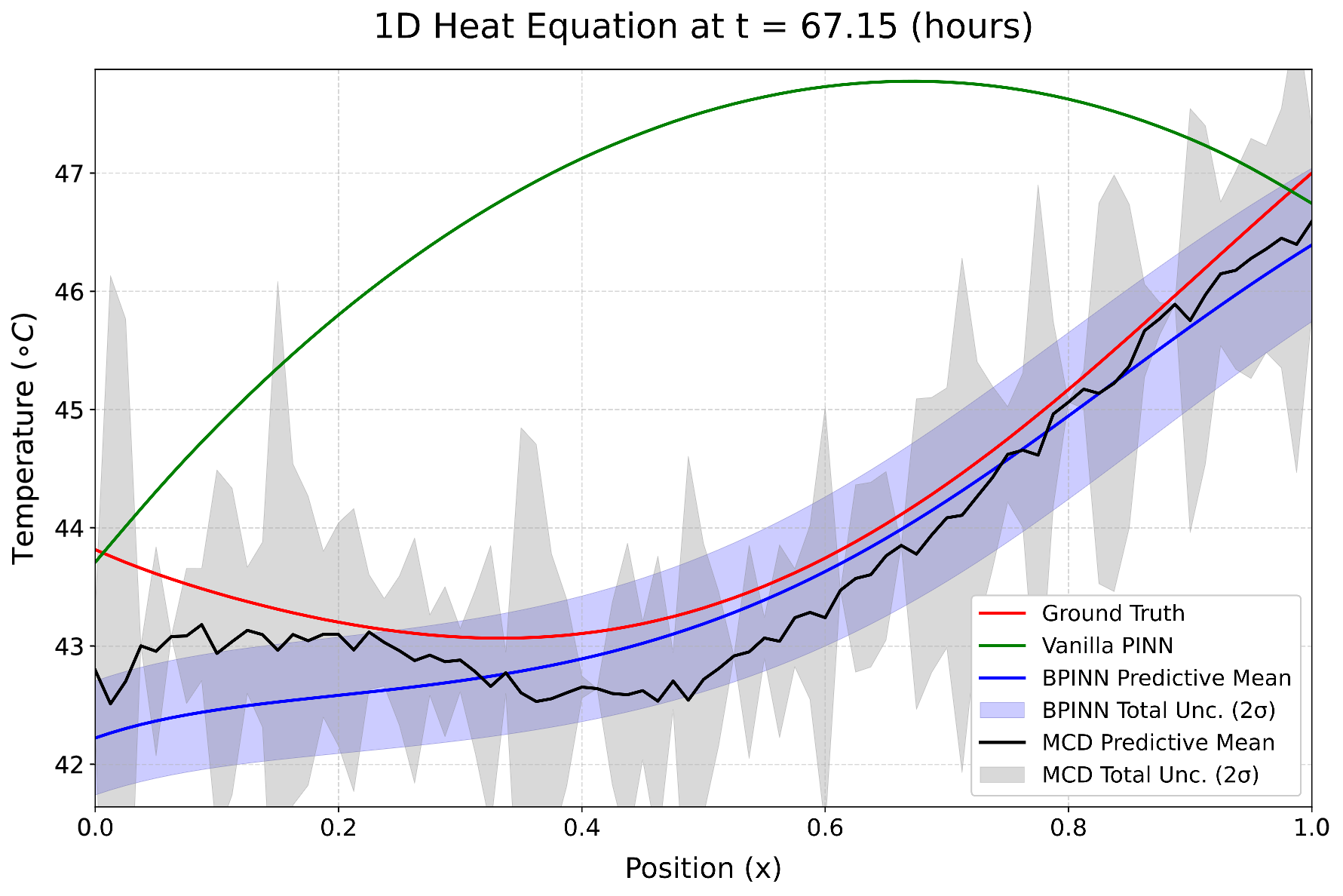}
	\includegraphics[width=0.33\columnwidth]{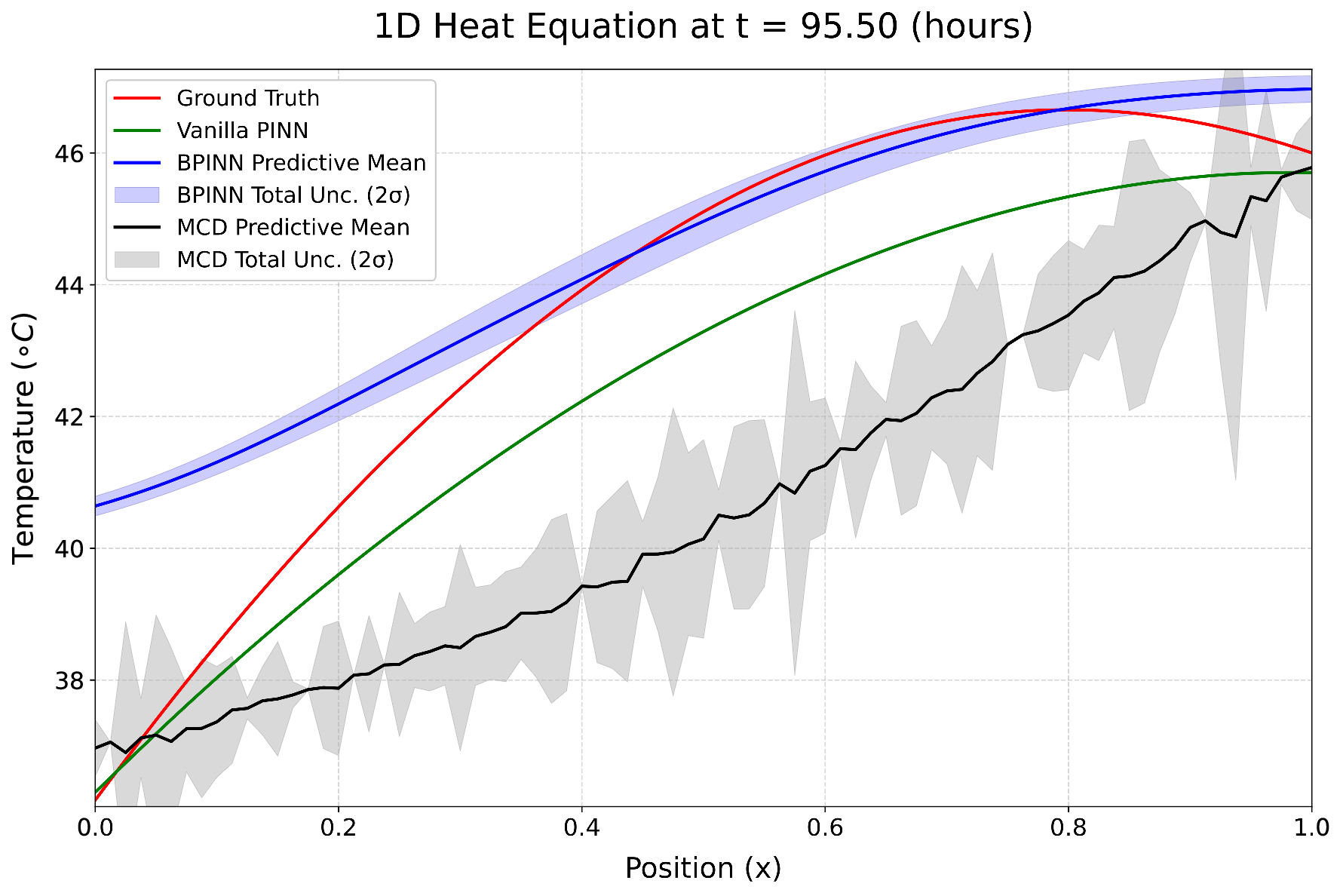}
	
	\caption{Spatiotemporal temperature estimates at different time instants for vanilla PINN, dropout-PINN, and Bayesian PINN (mean and standard deviation) models and ground truth values.}
	\label{fig:MeanTemperatureEstimation_Instants}
\end{figure*}

Figure~\ref{fig:MeanTemperatureEstimation_Instants} compares probabilistic spatiotemporal temperature estimates for the B-PINN, dropout-PINN, and vanilla PINN at selected time instants. The B-PINN consistently provides more accurate estimates than both alternatives. Importantly, the B-PINN not only delivers mean predictions, but also quantifies the uncertainty of the estimates. In most cases, the ground truth falls within the confidence intervals, in line with the PICP metric. The uncertainty bands are adaptive, reflecting prediction confidence that varies across the spatiotemporal domain. In contrast, the dropout-PINN produces wider and less consistent uncertainty bands, which would lead to overly conservative decision-making.

\begin{figure}[!htb]
	\centering
	\subfloat[]{\includegraphics[width=0.45\columnwidth]{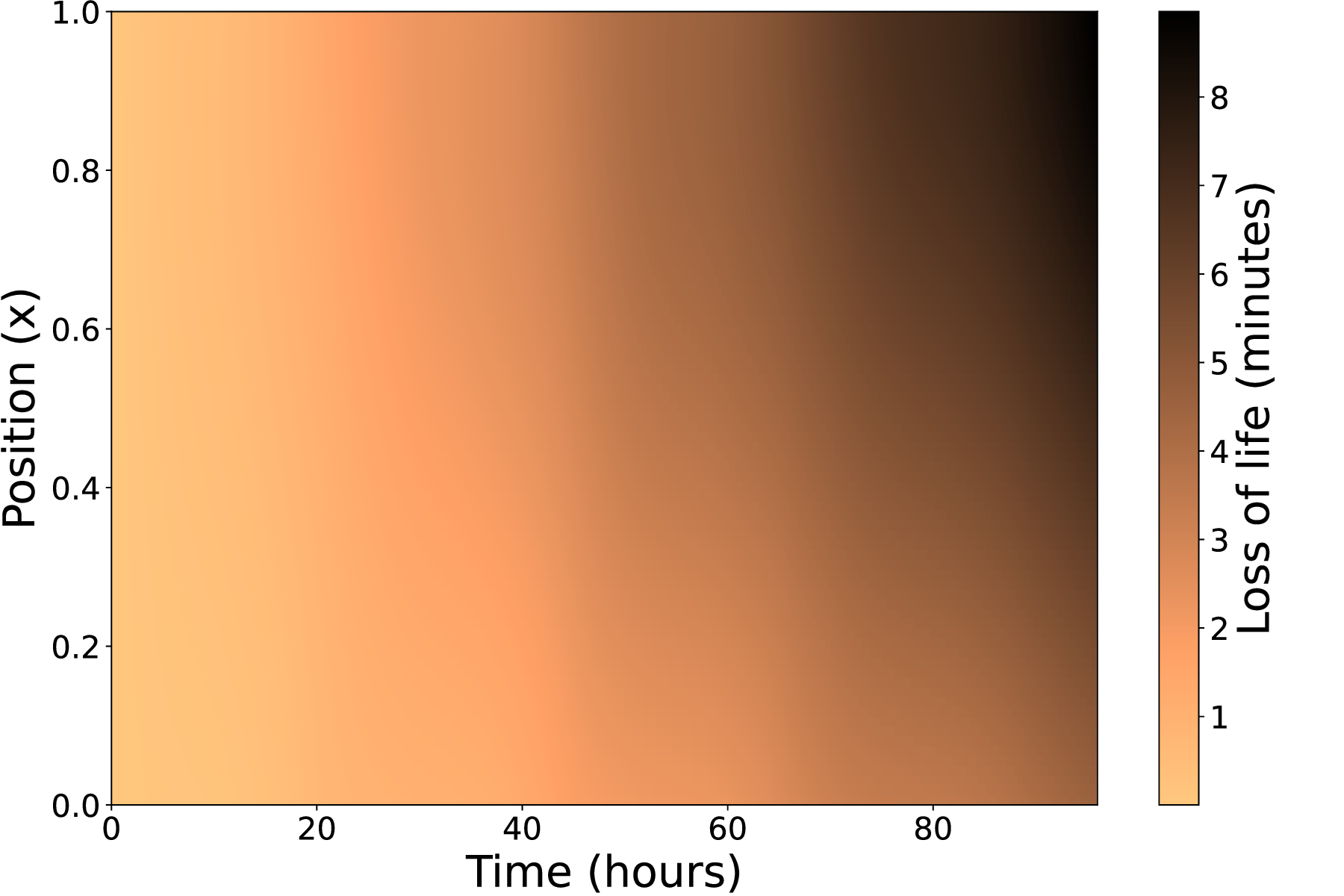}\label{subfig:1}}
	\hspace{0.02\textwidth}
	\subfloat[]{\includegraphics[width=0.45\columnwidth]{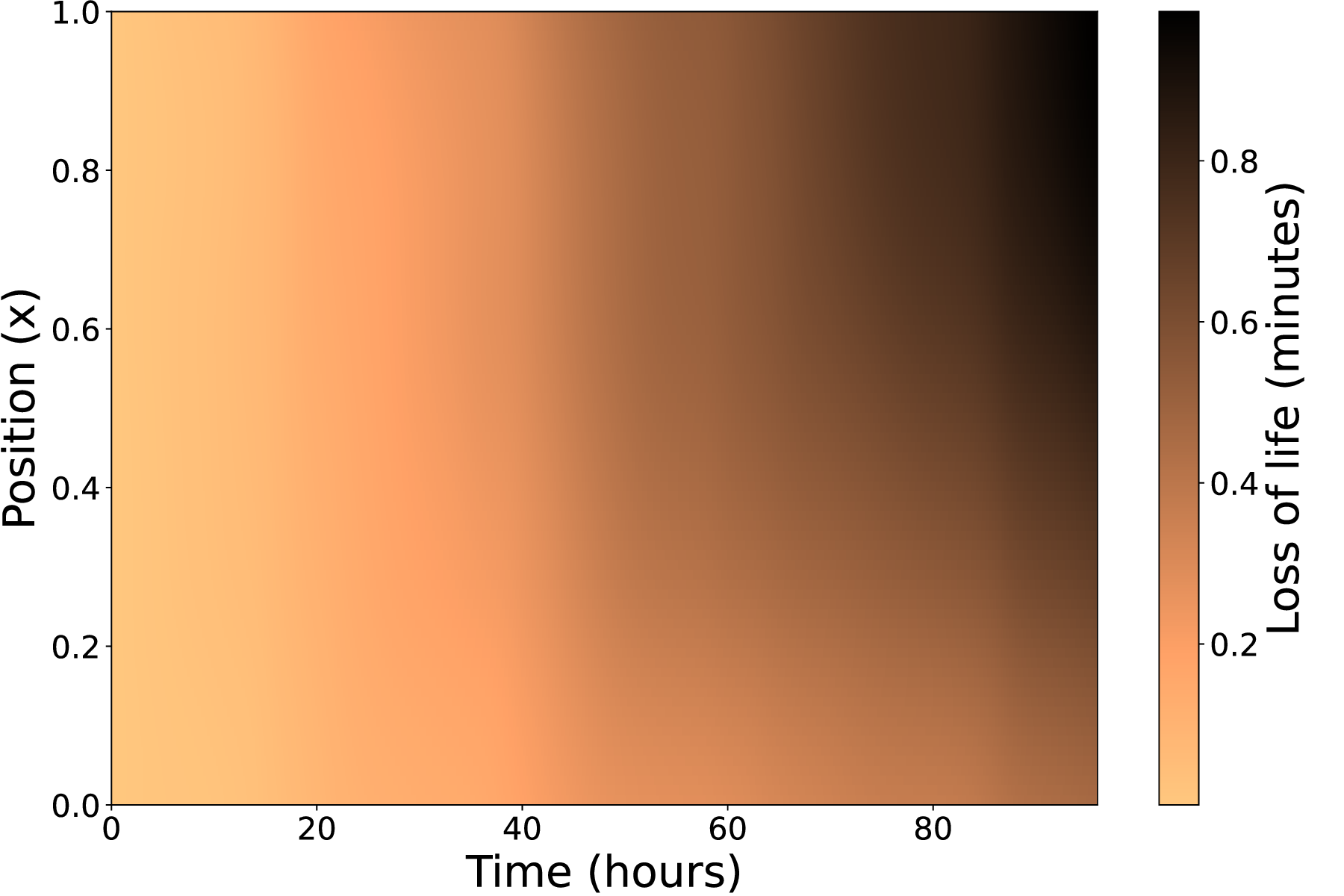}\label{subfig:2}}
	\caption{Probabilistic spatiotemporal transformer ageing $p(\hat{V}(x,t))$: (a) mean values (b) standard deviation.}
	\label{fig:MeanAgeingEstimation}
\end{figure}

Building on the probabilistic temperature field, Figure~\ref{fig:MeanAgeingEstimation} shows the spatiotemporal ageing estimates $p(\hat{V}(x,t))$ obtained with the B-PINN, including both the mean and standard deviation. After 4 days of operation under the loading profile of Figure~\ref{fig:AvailableTimeSeries}, the maximum transformer loss-of-life (cf. Eq.~(\ref{eq:LoL})) is 8.964 minutes according to the B-PINN mean estimate (Figure~\ref{fig:MeanAgeingEstimation}(a)), with a maximum deviation of 0.985 minutes (Figure~\ref{fig:MeanAgeingEstimation}(b)). This probabilistic ageing estimation framework supports informed maintenance decisions by balancing worst-case scenarios with less conservative, risk-adjusted strategies.



\begin{figure*}[!htb]
	\centering
	\subfloat[]{\includegraphics[width=0.3\columnwidth]{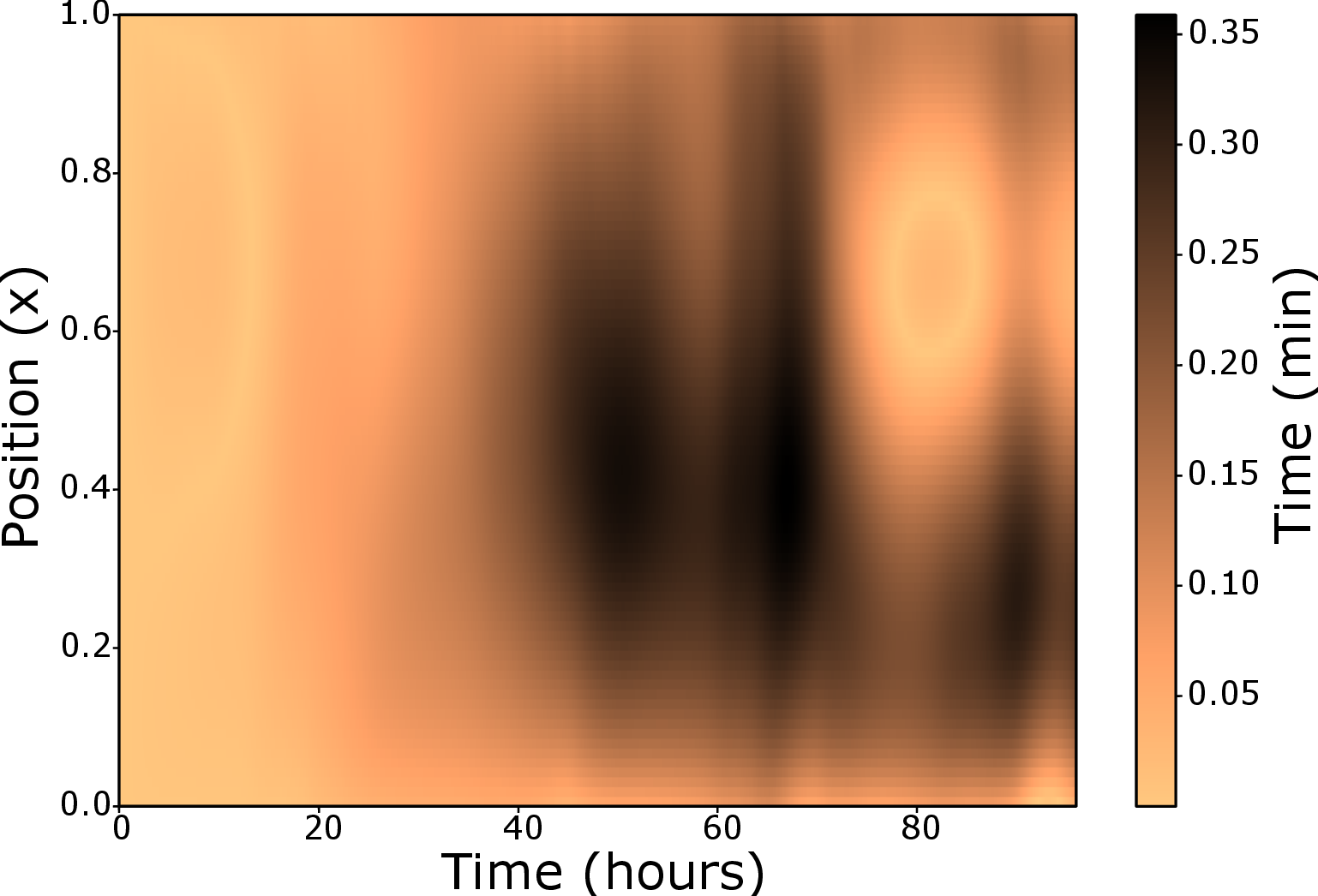}\label{subfig:3}}
	\hspace{0.02\textwidth}
	\subfloat[]{\includegraphics[width=0.3\columnwidth]{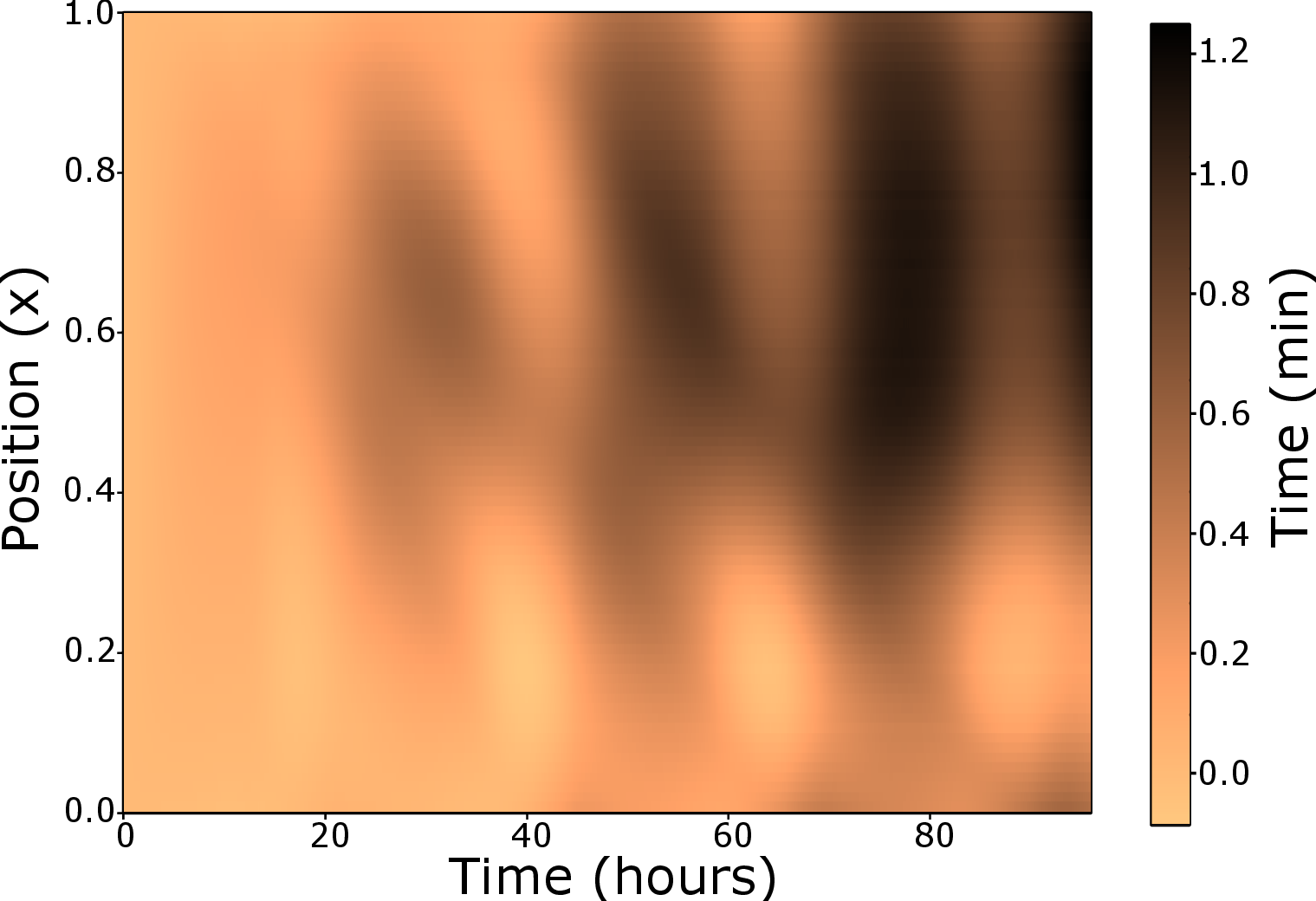}\label{subfig:4}}
	\hspace{0.02\textwidth}
	\subfloat[]{\includegraphics[width=0.3\columnwidth]{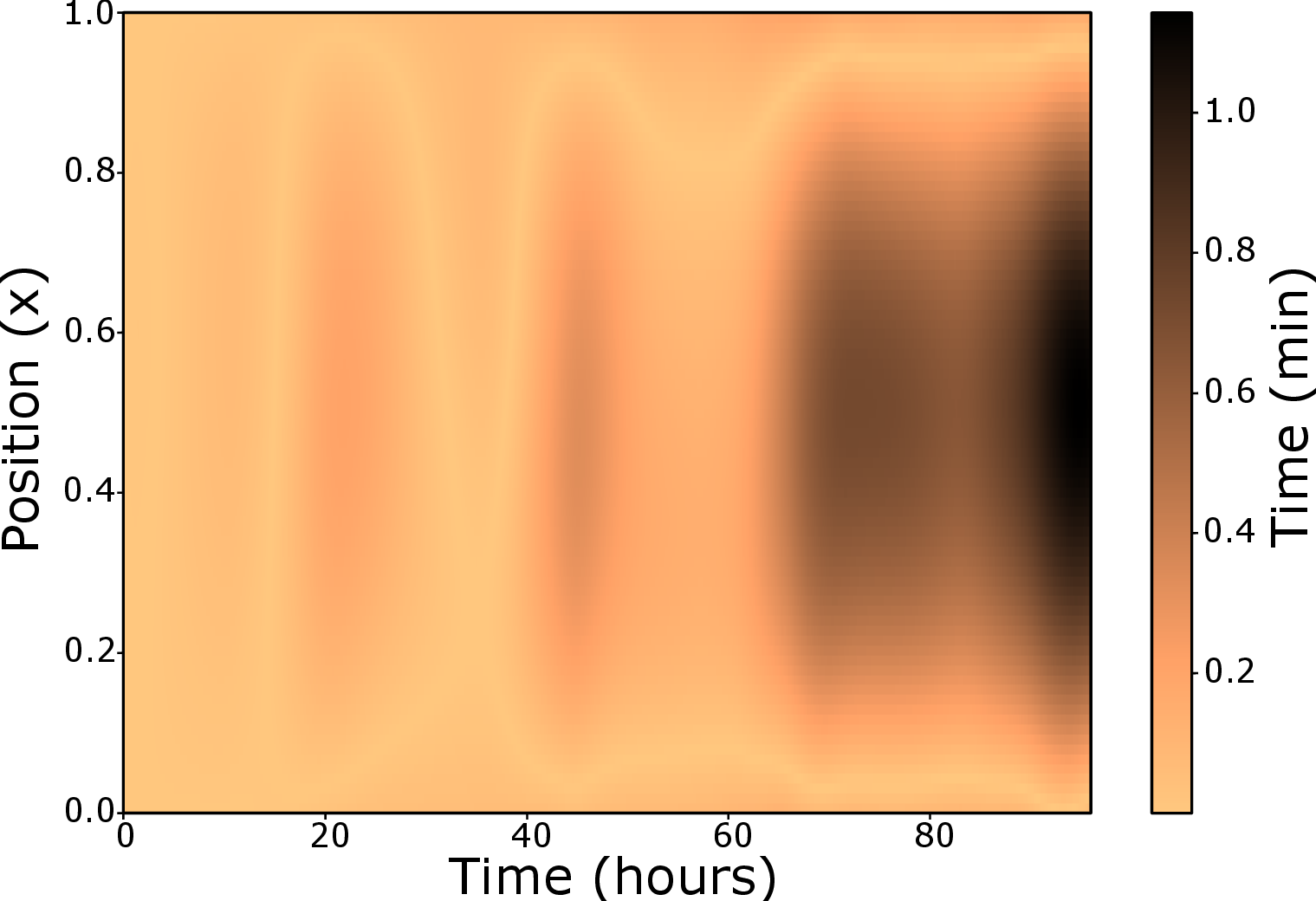}\label{subfig:temp_error_std_PINN}}
	\caption{Transformer ageing estimation error for (a) mean B-PINN, (b) mean dropout-PINN, and (c) vanilla PINN.}
	\label{fig:MeanAgeingEstimationError_PINN}
\end{figure*}

The ageing estimation errors of different models were further evaluated relative to the FEM-based ageing estimates. Figure~\ref{fig:MeanAgeingEstimationError_PINN} shows the results for vanilla PINN, dropout-PINN, and B-PINN models. The B-PINN achieved the lowest error, with a worst-case ageing estimation error of 0.35 minutes (4.06\%) (Figure~\ref{fig:MeanAgeingEstimationError_PINN}(a)). The dropout-PINN and vanilla PINN achieved worst-case errors of 1.25 minutes (14.22\%) and 1.28 minutes (14.57\%), respectively (Figure~\ref{fig:MeanAgeingEstimationError_PINN}(b)). The small standard deviation of the B-PINN ageing estimates (Figure~\ref{fig:MeanAgeingEstimation}(b)) further indicates that prediction variability is limited, avoiding the need for conservative decisions.

Overall, the probabilistic spatiotemporal ageing estimates produced by the B-PINN provide a more informative basis for asset health monitoring. On one hand, the spatiotemporal ageing maps enable maintenance planning that accounts for localized ageing effects, improving conventional practices based on worst-case, spatially-agnostic estimates. On the other hand, the explicit quantification of predictive uncertainty allows operators to assess reliability and adapt decisions accordingly, ranging from conservative interventions in uncertain scenarios to optimized actions in reliable regimes.


\subsection{Robustness to Noise}
\label{ss:Sensitivity}

After observing the probabilistic prediction results, the effect of different noise levels on the best-performing model (\textit{i.e.} B-PINN) will be examined. Namely, noise was introduced in the initial condition samples ($\sigma_i$) and residual samples ($\sigma_r$), and the resulting performance was evaluated under three prior configurations (Gaussian, spike-and-slab, and Laplace), while keeping all other base-case configuration parameters fixed (cf. Subsection~\ref{ss:Experiments}). The results are summarized in Table~\ref{table:UQ_noise}.

\begin{table*}[!htb]
	\small
	\centering
	\caption{Comparison of different noise levels and prior distributions for different UQ metrics with remaining parameters fixed  (2 layers, 50 neurons, $N_i$=200, $N_r$=10000). Best results highlighted.}
	\setlength{\tabcolsep}{0.7pt}
	\resizebox{\columnwidth}{!}{
	\begin{tabular}{|c|ccc|ccc|ccc|}
		\hline
		\multirow{2}{*}{\textbf{Noise Level}} & \multicolumn{3}{c|}{\textbf{Gaussian}} & \multicolumn{3}{c|}{\textbf{Spike-and-Slab}} & \multicolumn{3}{c|}{\textbf{Laplace}} \\
		\cline{2-10}
		& PICP ($\uparrow$) & CRPS ($\downarrow$) & NLL($\downarrow$) & PICP ($\uparrow$) & CRPS($\downarrow$) & NLL ($\downarrow$) & PICP ($\uparrow$) & CRPS ($\downarrow$) & NLL ($\downarrow$) \\
		\hline
		$\sigma_i$=$\sigma_r$=$\mathcal{N}(0,\!0.01^2)$ & \cellcolor{selected}0.416{\footnotesize $\pm$.041} & 0.092{\footnotesize $\pm$.026} & -0.374{\footnotesize$\pm$.218} & 0.348{\footnotesize $\pm$.083} & 0.131{\footnotesize $\pm$.045} & 0.127{\footnotesize $\pm$.733} &  0.324 {\footnotesize $\pm$.034}& \cellcolor{selected}0.077{\footnotesize $\pm$.009} & \cellcolor{selected}-0.478{\footnotesize$\pm$.071}\\
		$\sigma_i$=$\sigma_r$=$\mathcal{N}(0,\!0.05^2)$ & 0.242 {\footnotesize $\pm$.029} & 0.151{\footnotesize $\pm$.050} & 0.192{\footnotesize $\pm$.237} & 0.266{\footnotesize $\pm$.048} & 0.225{\footnotesize $\pm$.024} & 0.482{\footnotesize $\pm$.161} & 0.245{\footnotesize $\pm$.025} & 0.179{\footnotesize $\pm$.014} & 0.285{\footnotesize $\pm$.105}  \\
		$\sigma_i$=$\sigma_r$=$\mathcal{N}(0,\!0.1^2)$ & 0.191{\footnotesize $\pm$.037} & 0.179{\footnotesize $\pm$.041} & 0.351{\footnotesize $\pm$.125} & 0.215{\footnotesize $\pm$.026} & 0.275{\footnotesize $\pm$.097} & 0.957{\footnotesize $\pm$.805} & 0.216{\footnotesize $\pm$.043} & 0.202{\footnotesize $\pm$.011} & 0.441{\footnotesize $\pm$.050} \\
		\hline
	\end{tabular} }
	\label{table:UQ_noise}
\end{table*}

The results indicate a deterioration in probabilistic performance as the noise level increases. Higher noise reduces the PICP values, indicating a smaller proportion of true values fall within the predicted intervals, which compromises the reliability of the uncertainty estimates. At the same time, CRPS and NLL increase with noise, indicating a decline in the quality and calibration of the predictive distributions. Specifically, higher CRPS indicates less accurate probabilistic forecasts, while higher NLL values show that the model assigns lower likelihood to observed outcomes.


These trends are consistent across all priors, but their magnitudes are different. The Gaussian prior achieves the highest PICP values, but simultaneously produces worse NLL scores compared to the Laplace prior. This suggests that the Gaussian prior generates wider predictive intervals, capturing more true values but at the expense of less accurate probability assignment. Conversely, the Laplace prior provides better calibration (lower NLL and CRPS) despite offering narrower intervals and lower coverage. This illustrates the trade-off between interval coverage (PICP) and calibration quality (NLL, CRPS), showing that a higher PICP may not imply superior probabilistic performance.

Overall, the results confirm that increasing noise in both initial and residual data degrades the probabilistic performance and calibration of the B-PINN model. These findings emphasize the importance of carefully accounting for data noise and quality in UQ tasks.


\subsection{Computational Cost}
\label{ss:ComputationalCost}

Table~\ref{table:ComputationalCost} reports the computational cost of the different models, obtained on an AMD 4900HS processor with 32GB of RAM for 466560 evaluation points (81 in space, 5760 in time). The dropout-PINN achieves the shortest training time because randomly deactivating neurons reduces the effective model size during optimization. In contrast, the vanilla PINN requires more training time, and the B-PINN incurs additional cost due to uncertainty quantification. The FEM does not require training.

\begin{table}[!htb]
	\centering
	\small
	\caption{Computational cost of the evaluated models. Train and evaluation times averaged over repeated runs.}
	\setlength{\tabcolsep}{9.5pt}
	\begin{tabular}{|c|c| c|} \hline
		\textbf{Model} &  \textbf{Train [s]} & \textbf{Evaluation [s]}  \\ \hline 
		Vanilla-PINN  & 719.62 \footnotesize{$\pm$ 8.64} & 3.73 \footnotesize{$\pm$ 0.05} \\ \cline{1-1}
		Dropout-PINN  & 685.66 \footnotesize{$\pm$4.06} & 110.22\footnotesize{$\pm$ 0.64} \\ \cline{1-1}
		B-PINN    & 785.72\footnotesize{$\pm$ 24.12} & 498.38\footnotesize{$\pm$ 2.87} \\ \hline
		FEM  & \multicolumn{2}{|c|}{25.83  \footnotesize{$\pm$ 3.69}} \\ \hline
	\end{tabular}
	\label{table:ComputationalCost}
\end{table}

During evaluation, the vanilla PINN serves as the baseline without UQ and therefore avoids additional computational overhead. The dropout-PINN requires multiple stochastic forward passes to estimate predictive distributions, leading to higher evaluation time than vanilla PINN, but lower than B-PINN. The B-PINN incurs the highest evaluation cost because posterior inference requires repeated sampling across weight distributions. The FEM is the most efficient when training and evaluation are considered jointly. However, for more complex PDEs, the computational cost of FEM increases rapidly and PINNs provide a scalable and flexible alternative.

\section{Conclusions}
\label{sec:Conclusions}

Scientific Machine Learning (SciML) combines physical knowledge with machine learning (ML), resulting in predictive models that generalize better than purely data-driven approaches and require less data. Hybrid Prognostics and Health Management (PHM) methods share similarities with SciML, as both integrate physics-of-failure models and data. However, SciML provides a more principled learning framework that, to the best of the authors’ knowledge, has not been fully explored within PHM.

This work introduces a Bayesian Physics-Informed Neural Network (B-PINN) framework for probabilistic ageing estimation in electrical transformers. The method integrates the heat diffusion partial differential equation with transformer loading and temperature data, while modeling neural network weights probabilistically to capture epistemic uncertainty. Several prior distributions (Gaussian, spike-and-slab, Laplace) were evaluated, along with a range of B-PINN configurations (varying network width, number of initial and residual samples, and noise levels), and a dropout-PINN baseline for comparison.

Numerical results demonstrate that B-PINNs improve reliability over deterministic PINNs by providing both accurate predictions and well-calibrated uncertainty estimates. Compared with dropout-PINNs, B-PINNs yield more consistent and sharper predictive intervals, reducing the tendency toward overly conservative estimates. Among the priors, the Laplace prior consistently delivers the best performance across different settings, while the Gaussian prior performs robustly under high noise conditions. The spike-and-slab prior, in contrast, is more sensitive to data quality. Sensitivity analysis indicates that noise levels and the number of initial condition samples have the greatest influence on uncertainty quantification, whereas the number of residual samples has a comparatively lower influence.

Beyond methodological contributions, the proposed B-PINN framework offers practical benefits for transformer asset management. By quantifying and propagating epistemic uncertainty, the method supports informed decision-making under data scarcity, mitigating the risks of overly conservative or unsafe maintenance strategies.

The present implementation employs variational inference for posterior estimation. Future research will explore alternative Bayesian inference methods and physics-informed priors to further enhance uncertainty quantification in SciML-based PHM frameworks.

\section*{Acknowledgements}

This work has been partially funded by the Spanish State Research Agency (grant No. PID2024-156284OA-I00) and the Basque Government (grants IT1504-22, KK-2024/00030). Jose I. Aizpurua is funded by the Ramón y Cajal Fellowship, Spanish State Research Agency (grant No. RYC2022-037300-I), co-funded by MCIU/AEI/10.13039/501100011033 and FSE+.


\bibliographystyle{unsrtnat}
\bibliography{references}  

\begin{thebibliography}{51}
\providecommand{\natexlab}[1]{#1}
\providecommand{\url}[1]{\texttt{#1}}
\expandafter\ifx\csname urlstyle\endcsname\relax
  \providecommand{\doi}[1]{doi: #1}\else
  \providecommand{\doi}{doi: \begingroup \urlstyle{rm}\Url}\fi

\bibitem[Karniadakis et~al.(2021)Karniadakis, Kevrekidis, Lu, Perdikaris, Wang, and Yang]{karniadakis2021physicsinformed}
George~Em Karniadakis, Ioannis~G. Kevrekidis, Lu~Lu, Paris Perdikaris, Sifan Wang, and Liu Yang.
\newblock Physics-informed machine learning.
\newblock \emph{Nature Reviews Physics}, 3\penalty0 (6):\penalty0 422--440, 2021.
\newblock ISSN 25225820.
\newblock \doi{10.1038/s42254-021-00314-5}.

\bibitem[Raissi et~al.(2019{\natexlab{a}})Raissi, Perdikaris, and Karniadakis]{Raissi_19}
M.~Raissi, P.~Perdikaris, and G.E. Karniadakis.
\newblock Physics-informed neural networks: A deep learning framework for solving forward and inverse problems involving nonlinear partial differential equations.
\newblock \emph{Journal of Computational Physics}, 378:\penalty0 686--707, 2019{\natexlab{a}}.
\newblock ISSN 0021-9991.
\newblock \doi{10.1016/j.jcp.2018.10.045}.

\bibitem[Liu et~al.(2025)Liu, Wang, Vaidya, Ruehle, Halverson, Soljačić, Hou, and Tegmark]{KAN_ICLR}
Ziming Liu, Yixuan Wang, Sachin Vaidya, Fabian Ruehle, James Halverson, Marin Soljačić, Thomas~Y. Hou, and Max Tegmark.
\newblock Kan: Kolmogorov-arnold networks.
\newblock In \emph{{International Conference on Learning Representations (ICLR)}}, 2025.
\newblock \doi{https://doi.org/10.48550/arXiv.2404.19756}.

\bibitem[Kovachki et~al.(2023)Kovachki, Li, Liu, Azizzadenesheli, Bhattacharya, Stuart, and Anandkumar]{NeuralOperators_23}
Nikola Kovachki, Zongyi Li, Burigede Liu, Kamyar Azizzadenesheli, Kaushik Bhattacharya, Andrew Stuart, and Anima Anandkumar.
\newblock Neural operator: learning maps between function spaces with applications to pdes.
\newblock \emph{J. Mach. Learn. Res.}, 24\penalty0 (1), January 2023.
\newblock ISSN 1532-4435.
\newblock \doi{10.5555/3648699.3648788}.

\bibitem[Li et~al.(2024{\natexlab{a}})Li, Zheng, Kovachki, Jin, Chen, Liu, Azizzadenesheli, and Anandkumar]{PINO_24}
Zongyi Li, Hongkai Zheng, Nikola Kovachki, David Jin, Haoxuan Chen, Burigede Liu, Kamyar Azizzadenesheli, and Anima Anandkumar.
\newblock Physics-informed neural operator for learning partial differential equations.
\newblock \emph{ACM / IMS J. Data Sci.}, 1\penalty0 (3), May 2024{\natexlab{a}}.
\newblock \doi{10.1145/3648506}.

\bibitem[Toscano et~al.(2025)Toscano, Oommen, Varghese, Zou, Daryakenari, Wu, and Karniadakis]{toscano2025}
Juan~Diego Toscano, Vivek Oommen, Alan~John Varghese, Zongren Zou, Nazanin~Ahmadi Daryakenari, Chenxi Wu, and George~Em Karniadakis.
\newblock From pinns to pikans: Recent advances in physics-informed machine learning.
\newblock \emph{Mach. Learn. Comput. Sci. Eng.}, 1\penalty0 (15), 2025.
\newblock \doi{10.1007/s44379-025-00015-1}.

\bibitem[Aizpurua and Catterson(2015)]{aizpurua2015towards}
Jose~Ignacio Aizpurua and Victoria~M Catterson.
\newblock Towards a methodology for design of prognostic systems.
\newblock In \emph{{Annual Conference of the PHM Society}}, volume~7, 2015.
\newblock \doi{10.36001/phmconf.2015.v7i1.2646}.

\bibitem[Guo et~al.(2020)Guo, Li, and Li]{Guo_20}
Jian Guo, Zhaojun Li, and Meiyan Li.
\newblock A review on prognostics methods for engineering systems.
\newblock \emph{IEEE Transactions on Reliability}, 69\penalty0 (3):\penalty0 1110--1129, 2020.
\newblock \doi{10.1109/TR.2019.2957965}.

\bibitem[Zio(2022)]{Zio_22}
Enrico Zio.
\newblock Prognostics and health management (phm): Where are we and where do we (need to) go in theory and practice.
\newblock \emph{Reliability Engineering \& System Safety}, 218:\penalty0 108119, 2022.
\newblock ISSN 0951-8320.
\newblock \doi{10.1016/j.ress.2021.108119}.

\bibitem[Xu et~al.(2023)Xu, Kohtz, Boakye, Gardoni, and Wang]{Xu_23}
Yanwen Xu, Sara Kohtz, Jessica Boakye, Paolo Gardoni, and Pingfeng Wang.
\newblock Physics-informed machine learning for reliability and systems safety applications: State of the art and challenges.
\newblock \emph{Reliability Engineering \& System Safety}, 230:\penalty0 108900, 2023.
\newblock ISSN 0951-8320.
\newblock \doi{10.1016/j.ress.2022.108900}.

\bibitem[{Arias Chao} et~al.(2022){Arias Chao}, Kulkarni, Goebel, and Fink]{Chao22}
Manuel {Arias Chao}, Chetan Kulkarni, Kai Goebel, and Olga Fink.
\newblock Fusing physics-based and deep learning models for prognostics.
\newblock \emph{Reliability Engineering \& System Safety}, 217:\penalty0 107961, 2022.
\newblock ISSN 0951-8320.
\newblock \doi{10.1016/j.ress.2021.107961}.

\bibitem[Daigle et~al.(2015)Daigle, Roychoudhury, and Bregon]{daigle2015model}
Matthew Daigle, Indranil Roychoudhury, and Anibal Bregon.
\newblock Model-based prognostics of hybrid systems.
\newblock In \emph{{Annual Conference of the PHM Society}}, volume~7, 2015.
\newblock \doi{10.36001/phmconf.2015.v7i1.2586}.

\bibitem[Alcibar et~al.(2025)Alcibar, Aizpurua, Zugasti, and Peñagarikano]{Alcibar_2025}
Jokin Alcibar, Jose~I. Aizpurua, Ekhi Zugasti, and Oier Peñagarikano.
\newblock A hybrid probabilistic battery health management approach for robust inspection drone operations.
\newblock \emph{Engineering Applications of Artificial Intelligence}, 146:\penalty0 110246, 2025.
\newblock ISSN 0952-1976.
\newblock \doi{10.1016/j.engappai.2025.110246}.

\bibitem[Wang et~al.(2024)Wang, Zhai, Zhao, Di, and Chen]{wang2024physics}
Fujin Wang, Zhi Zhai, Zhibin Zhao, Yi~Di, and Xuefeng Chen.
\newblock Physics-informed neural network for lithium-ion battery degradation stable modeling and prognosis.
\newblock \emph{Nature Communications}, 15\penalty0 (1):\penalty0 4332, 2024.
\newblock \doi{https://doi.org/10.1038/s41467-024-48779-z}.

\bibitem[Ramirez et~al.(2025)Ramirez, Pino, Pardo, Sanz, {del Rio}, Ortiz, Morozovska, and Aizpurua]{Ramirez_25}
Ibai Ramirez, Joel Pino, David Pardo, Mikel Sanz, Luis {del Rio}, Alvaro Ortiz, Kateryna Morozovska, and Jose~I. Aizpurua.
\newblock Residual-based attention physics-informed neural networks for spatio-temporal ageing assessment of transformers operated in renewable power plants.
\newblock \emph{Engineering Applications of Artificial Intelligence}, 139:\penalty0 109556, 2025.
\newblock ISSN 0952-1976.
\newblock \doi{10.1016/j.engappai.2024.109556}.

\bibitem[Nascimento et~al.(2023)Nascimento, Viana, Corbetta, and Kulkarni]{nascimento2023framework}
Renato~G Nascimento, Felipe~AC Viana, Matteo Corbetta, and Chetan~S Kulkarni.
\newblock A framework for li-ion battery prognosis based on hybrid bayesian physics-informed neural networks.
\newblock \emph{Scientific Reports}, 13\penalty0 (1):\penalty0 13856, 2023.
\newblock \doi{10.1038/s41598-023-33018-0}.

\bibitem[Fernández et~al.(2023)Fernández, Chiachío, Chiachío, Barros, and Corbetta]{Juan_2023}
Juan Fernández, Juan Chiachío, Manuel Chiachío, José Barros, and Matteo Corbetta.
\newblock Physics-guided bayesian neural networks by abc-ss: Application to reinforced concrete columns.
\newblock \emph{Engineering Applications of Artificial Intelligence}, 119:\penalty0 105790, 2023.
\newblock ISSN 0952-1976.
\newblock \doi{10.1016/j.engappai.2022.105790}.

\bibitem[Sankararaman(2015)]{sankararaman2015significance}
Shankar Sankararaman.
\newblock Significance, interpretation, and quantification of uncertainty in prognostics and remaining useful life prediction.
\newblock \emph{Mechanical Systems and Signal Processing}, 52:\penalty0 228--247, 2015.

\bibitem[Biggio et~al.(2021)Biggio, Wieland, Chao, Kastanis, and Fink]{biggio2021uncertainty}
Luca Biggio, Alexander Wieland, Manuel~Arias Chao, Iason Kastanis, and Olga Fink.
\newblock Uncertainty-aware prognosis via deep gaussian process.
\newblock \emph{IEEE Access}, 9:\penalty0 123517--123527, 2021.

\bibitem[Javanmardi and H{\"u}llermeier(2023)]{javanmardi2023conformal}
Alireza Javanmardi and Eyke H{\"u}llermeier.
\newblock Conformal prediction intervals for remaining useful lifetime estimation.
\newblock \emph{International Journal of Prognostics and Health Management}, 14\penalty0 (2), 2023.

\bibitem[Alcibar et~al.(2024)Alcibar, Aizpurua, and Zugasti]{PHME_Jokin}
Jokin Alcibar, Jose~Ignacio Aizpurua, and Ekhi Zugasti.
\newblock Towards a probabilistic fusion approach for robust battery prognostics.
\newblock In \emph{{PHM Society European Conference}}, volume~8, 2024.
\newblock \doi{10.36001/phme.2024.v8i1.3981}.

\bibitem[Nemani et~al.(2023)Nemani, Biggio, Huan, Hu, Fink, Tran, Wang, Zhang, and Hu]{nemani2023uncertainty}
Venkat Nemani, Luca Biggio, Xun Huan, Zhen Hu, Olga Fink, Anh Tran, Yan Wang, Xiaoge Zhang, and Chao Hu.
\newblock Uncertainty quantification in machine learning for engineering design and health prognostics: A tutorial.
\newblock \emph{Mechanical Systems and Signal Processing}, 205:\penalty0 110796, 2023.

\bibitem[Basora et~al.(2025)Basora, Viens, Chao, and Olive]{basora2025benchmark}
Luis Basora, Arthur Viens, Manuel~Arias Chao, and Xavier Olive.
\newblock A benchmark on uncertainty quantification for deep learning prognostics.
\newblock \emph{Reliability Engineering \& System Safety}, 253:\penalty0 110513, 2025.

\bibitem[Linka et~al.(2022)Linka, Schäfer, Meng, Zou, Karniadakis, and Kuhl]{Linka2022}
Kevin Linka, Amelie Schäfer, Xuhui Meng, Zongren Zou, George~Em Karniadakis, and Ellen Kuhl.
\newblock Bayesian physics informed neural networks for real-world nonlinear dynamical systems.
\newblock \emph{Computer Methods in Applied Mechanics and Engineering}, 402:\penalty0 115346, 2022.
\newblock ISSN 0045-7825.
\newblock \doi{10.1016/j.cma.2022.115346}.

\bibitem[Flores et~al.(2025)Flores, Graf, Protopapas, and Pichara]{flores2025}
Pablo Flores, Olga Graf, Pavlos Protopapas, and Karim Pichara.
\newblock Improved uncertainty quantification in physics-informed neural networks using error bounds and solution bundles.
\newblock In \emph{{The Conference on Uncertainty in Artificial Intelligence}}, 2025.
\newblock \doi{10.48550/arXiv.2505.06459}.

\bibitem[Garg and Chakraborty(2023)]{Garg2023}
Shailesh Garg and Souvik Chakraborty.
\newblock Vb-deeponet: A bayesian operator learning framework for uncertainty quantification.
\newblock \emph{Engineering Applications of Artificial Intelligence}, 118:\penalty0 105685, 2023.
\newblock ISSN 0952-1976.
\newblock \doi{10.1016/j.engappai.2022.105685}.

\bibitem[Gao and Karniadakis(2025)]{gao2025scalable}
Zhiwei Gao and George~Em Karniadakis.
\newblock Scalable bayesian physics-informed kolmogorov-arnold networks.
\newblock \emph{arXiv preprint arXiv:2501.08501}, 2025.
\newblock \doi{10.48550/arXiv.2501.08501}.

\bibitem[Li et~al.(2024{\natexlab{b}})Li, Long, Deng, Jiang, Zhou, Jiang, and Zhang]{Li_2024}
Jinwu Li, Xiangyun Long, Xinyang Deng, Wen Jiang, Kai Zhou, Chao Jiang, and Xiaoge Zhang.
\newblock A principled distance-aware uncertainty quantification approach for enhancing the reliability of physics-informed neural network.
\newblock \emph{Reliability Engineering \& System Safety}, 245:\penalty0 109963, 2024{\natexlab{b}}.
\newblock ISSN 0951-8320.
\newblock \doi{10.1016/j.ress.2024.109963}.

\bibitem[Psaros et~al.(2023)Psaros, Meng, Zou, Guo, and Karniadakis]{psaros2023uncertainty}
Apostolos~F Psaros, Xuhui Meng, Zongren Zou, Ling Guo, and George~Em Karniadakis.
\newblock Uncertainty quantification in scientific machine learning: Methods, metrics, and comparisons.
\newblock \emph{Journal of Computational Physics}, 477:\penalty0 111902, 2023.
\newblock \doi{10.1016/j.jcp.2022.111902}.

\bibitem[Nair et~al.(2025)Nair, Jacob, Howard, Drgona, and Stinis]{Epinet2025}
Ashish~S. Nair, Bruno Jacob, Amanda~A. Howard, Jan Drgona, and Panos Stinis.
\newblock E-pinns: Epistemic physics-informed neural networks, 2025.

\bibitem[Podina et~al.(2024)Podina, Rad, and Kohandel]{podina2024conformalized}
Lena Podina, Mahdi~Torabi Rad, and Mohammad Kohandel.
\newblock Conformalized physics-informed neural networks.
\newblock In \emph{{ICLR 2024 Workshop on AI4Differential Equations In Science}}, 2024.
\newblock \doi{10.48550/arXiv.2405.08111}.

\bibitem[Gal and Ghahramani(2016)]{Gal16}
Yarin Gal and Zoubin Ghahramani.
\newblock Dropout as a bayesian approximation: Representing model uncertainty in deep learning.
\newblock In Maria~Florina Balcan and Kilian~Q. Weinberger, editors, \emph{Proceedings of The 33rd International Conference on Machine Learning}, volume~48 of \emph{Proceedings of Machine Learning Research}, pages 1050--1059, New York, New York, USA, 20--22 Jun 2016. PMLR.
\newblock \doi{10.48550/arXiv.1506.02142}.

\bibitem[Lakshminarayanan et~al.(2017)Lakshminarayanan, Pritzel, and Blundell]{Lakshminarayanan_17}
Balaji Lakshminarayanan, Alexander Pritzel, and Charles Blundell.
\newblock Simple and scalable predictive uncertainty estimation using deep ensembles.
\newblock In \emph{{International Conference on Neural Information Processing Systems}}, NIPS'17, page 6405–6416, Red Hook, NY, USA, 2017. Curran Associates Inc.
\newblock ISBN 9781510860964.
\newblock \doi{10.48550/arXiv.1612.01474}.

\bibitem[Zhang et~al.(2019)Zhang, Lu, Guo, and Karniadakis]{Zhang_19}
Dongkun Zhang, Lu~Lu, Ling Guo, and George~Em Karniadakis.
\newblock Quantifying total uncertainty in physics-informed neural networks for solving forward and inverse stochastic problems.
\newblock \emph{Journal of Computational Physics}, 397:\penalty0 108850, 2019.
\newblock ISSN 0021-9991.
\newblock \doi{10.1016/j.jcp.2019.07.048}.

\bibitem[Jiang et~al.(2023)Jiang, Wang, Wen, Li, and Wang]{jiang2023practical}
Xinchao Jiang, Xin Wang, Ziming Wen, Enying Li, and Hu~Wang.
\newblock Practical uncertainty quantification for space-dependent inverse heat conduction problem via ensemble physics-informed neural networks.
\newblock \emph{International Communications in Heat and Mass Transfer}, 147:\penalty0 106940, 2023.
\newblock \doi{10.1016/j.icheatmasstransfer.2023.106940}.

\bibitem[Zou et~al.(2025)Zou, Wang, and Karniadakis]{zou2025learning}
Zongren Zou, Zhicheng Wang, and George~Em Karniadakis.
\newblock Learning and discovering multiple solutions using physics-informed neural networks with random initialization and deep ensemble.
\newblock \emph{arXiv preprint arXiv:2503.06320}, 2025.
\newblock \doi{10.48550/arXiv.2503.06320}.

\bibitem[Aizpurua et~al.(2023)Aizpurua, Ramirez, Lasa, Rio, Ortiz, and Stewart]{Aizpurua_23}
Jose~Ignacio Aizpurua, Ibai Ramirez, Iker Lasa, Luis~del Rio, Alvaro Ortiz, and Brian~G. Stewart.
\newblock Hybrid transformer prognostics framework for enhanced probabilistic predictions in renewable energy applications.
\newblock \emph{IEEE Trans. Pow. Del.}, 38\penalty0 (1):\penalty0 599--609, 2023.
\newblock \doi{10.1109/TPWRD.2022.3203873}.

\bibitem[{International Electrotechnical Commission}(2018)]{IEC60076_transf12}
{International Electrotechnical Commission}.
\newblock {Loading guide for oil-immersed power transformers}.
\newblock \emph{IEC 60076-7}, 2018.
\newblock URL \url{https://webstore.iec.ch/}.

\bibitem[Raissi et~al.(2019{\natexlab{b}})Raissi, Perdikaris, and Karniadakis]{Raisi_19}
M.~Raissi, P.~Perdikaris, and G.E. Karniadakis.
\newblock Physics-informed neural networks: A deep learning framework for solving forward and inverse problems involving nonlinear partial differential equations.
\newblock \emph{Journal of Computational Physics}, 378:\penalty0 686--707, 2019{\natexlab{b}}.
\newblock ISSN 0021-9991.
\newblock \doi{10.1016/j.jcp.2018.10.045}.

\bibitem[Wright et~al.(2022)Wright, Onodera, Stein, Wang, Schachter, Hu, and McMahon]{wright2022deep}
Logan~G Wright, Tatsuhiro Onodera, Martin~M Stein, Tianyu Wang, Darren~T Schachter, Zoey Hu, and Peter~L McMahon.
\newblock Deep physical neural networks trained with backpropagation.
\newblock \emph{Nature}, 601\penalty0 (7894):\penalty0 549--555, 2022.
\newblock \doi{10.1038/s41586-021-04223-6}.

\bibitem[Anagnostopoulos et~al.(2024)Anagnostopoulos, Toscano, Stergiopulos, and Karniadakis]{Agnostopoulos_RBA}
Sokratis~J. Anagnostopoulos, Juan~Diego Toscano, Nikolaos Stergiopulos, and George~Em Karniadakis.
\newblock Residual-based attention in physics-informed neural networks.
\newblock \emph{Computer Methods in Applied Mechanics and Engineering}, 421:\penalty0 116805, 2024.
\newblock ISSN 0045-7825.
\newblock \doi{10.1016/j.cma.2024.116805}.

\bibitem[Fortuin et~al.(2021)Fortuin, Garriga-Alonso, {van der Wilk}, and Aitchison]{Fortuin_21}
Vincent Fortuin, Adrià Garriga-Alonso, Mark {van der Wilk}, and Laurence Aitchison.
\newblock Bnnpriors: A library for bayesian neural network inference with different prior distributions.
\newblock \emph{Software Impacts}, 9:\penalty0 100079, 2021.
\newblock ISSN 2665-9638.
\newblock \doi{10.1016/j.simpa.2021.100079}.

\bibitem[Williams(1995)]{williams1995}
Peter~M. Williams.
\newblock Bayesian regularization and pruning using a laplace prior.
\newblock \emph{Neural Computation}, 7\penalty0 (1):\penalty0 117--143, 1995.
\newblock \doi{10.1162/neco.1995.7.1.117}.

\bibitem[Kendall and Gal(2017)]{kendall2017}
Alex Kendall and Yarin Gal.
\newblock What uncertainties do we need in bayesian deep learning for computer vision?
\newblock In I.~Guyon, U.~Von Luxburg, S.~Bengio, H.~Wallach, R.~Fergus, S.~Vishwanathan, and R.~Garnett, editors, \emph{Advances in Neural Information Processing Systems}, volume~30, pages 1--11. {Curran Associates, Inc.}, 2017.
\newblock \doi{10.48550/arXiv.1703.04977}.

\bibitem[Wang et~al.(2023)Wang, Sankaran, Wang, and Perdikaris]{wang2023expertsguidetrainingphysicsinformed}
Sifan Wang, Shyam Sankaran, Hanwen Wang, and Paris Perdikaris.
\newblock An expert's guide to training physics-informed neural networks, 2023.

\bibitem[Murphy(2012)]{murphy_2012}
Kevin~P Murphy.
\newblock \emph{Machine learning: a probabilistic perspective}.
\newblock MIT press, 2012.
\newblock ISBN 978-0262018029.

\bibitem[Bishop and Nasrabadi(2006)]{bishop_2006}
Christopher~M Bishop and Nasser~M Nasrabadi.
\newblock \emph{Pattern recognition and machine learning}, volume~4.
\newblock Springer, 2006.
\newblock \doi{10.1117/1.2819119}.

\bibitem[Zamo and Naveau(2018)]{zamo2018}
Micha{\"e}l Zamo and Philippe Naveau.
\newblock Estimation of the {{Continuous Ranked Probability Score}} with {{Limited Information}} and {{Applications}} to {{Ensemble Weather Forecasts}}.
\newblock \emph{Mathematical Geosciences}, 50\penalty0 (2):\penalty0 209--234, February 2018.
\newblock ISSN 1874-8953.
\newblock \doi{10.1007/s11004-017-9709-7}.

\bibitem[Gneiting et~al.(2005)Gneiting, Raftery, Westveld, and Goldman]{Gneiting2005}
Tilmann Gneiting, Adrian~E. Raftery, Anton~H. Westveld, and Tom Goldman.
\newblock Calibrated probabilistic forecasting using ensemble model output statistics and minimum {{CRPS}} estimation.
\newblock \emph{Monthly Weather Review}, 133\penalty0 (5):\penalty0 1098--1118, 2005.
\newblock \doi{10.1175/MWR2904.1}.

\bibitem[{Gonz{\'a}lez-Sope{\~n}a} et~al.(2021){Gonz{\'a}lez-Sope{\~n}a}, Pakrashi, and Ghosh]{GONZALEZ2021}
J.M. {Gonz{\'a}lez-Sope{\~n}a}, V.~Pakrashi, and B.~Ghosh.
\newblock An overview of performance evaluation metrics for short-term statistical wind power forecasting.
\newblock \emph{Renewable and Sustainable Energy Reviews}, 138:\penalty0 110515, 2021.
\newblock ISSN 1364-0321.
\newblock \doi{10.1016/j.rser.2020.110515}.

\bibitem[{The MathWorks, Inc.}(2023)]{matlabpdepe}
{The MathWorks, Inc.}
\newblock \emph{PDEPE Toolbox}.
\newblock Natick, Massachusetts, USA, 2023.
\newblock URL \url{https://mathworks.com/}.

\end{thebibliography}

\end{document}